%% file: paper.tex
\documentclass[10pt,twocolumn,letterpaper]{article}

\usepackage{cvpr}
\usepackage{times}
\usepackage{epsfig}
\usepackage{graphicx}
\usepackage{amsmath}
\usepackage{amssymb}
\usepackage{tabularx}

\usepackage{bm}
\usepackage{booktabs}

\usepackage[pagebackref=true,breaklinks=true,letterpaper=true,colorlinks,bookmarks=false]{hyperref}

\cvprfinalcopy %

\def\cvprPaperID{****} %

\begin{document}

\title{Self-supervised Learning of 3D Objects from Natural Images}

\author{
    Hiroharu Kato${}^\text{1}$ and Tatsuya Harada${}^\text{1,2}$\\
    ${}^\text{1}$The University of Tokyo, ${}^\text{2}$RIKEN\\
    {\tt\small \{kato,harada\}@mi.t.u-tokyo.ac.jp}
}

\maketitle
\thispagestyle{empty}

\begin{abstract}
\input{abstract.tex}
\end{abstract}

\input{1_introduction.tex}

\input{2_related_work.tex}
\input{3_method.tex}
\input{4_experiments.tex}

\input{5_conclusion.tex}
\input{6_acknowledgment.tex}

\clearpage
{
    \small
    \bibliographystyle{ieee_fullname}
    \bibliography{paper}
}

\clearpage
\appendix
\input{7_appendix.tex}

\end{document}

%% file: abstract.tex
We present a method to learn single-view reconstruction of the 3D shape, pose, and texture of objects from categorized natural images in a self-supervised manner. Since this is a severely ill-posed problem, carefully designing a training method and introducing constraints are essential. To avoid the difficulty of training all elements at the same time, we propose training category-specific base shapes with fixed pose distribution and simple textures first, and subsequently training poses and textures using the obtained shapes. Another difficulty is that shapes and backgrounds sometimes become excessively complicated to mistakenly reconstruct textures on object surfaces. To suppress it, we propose using strong regularization and constraints on object surfaces and background images. With these two techniques, we demonstrate that we can use natural image collections such as CIFAR-10 and PASCAL objects for training, which indicates the possibility to realize 3D object reconstruction on diverse object categories beyond synthetic datasets.

%% file: 1_introduction.tex
\vspace{-3mm}
\section{Introduction}
\label{sec:introduction}

By looking at an object at a glance, we humans can understand its 3D shape, orientation, and appearance on surfaces. Implementing this ability in machines, known as single-view 3D object reconstruction and object pose estimation in computer vision, has many practical applications such as robot grasping and augmented reality. Since this is a severely ill-posed problem, learning and leveraging the prior knowledge of objects is the key to this task.

Most works in this field use ShapeNet~\cite{chang2015shapenet}, a large-scale 3D CAD dataset, for training. Though recent technical advancement has realized to generate a high-quality 3D model from an image in object categories that are contained in ShapeNet~\cite{choy20163d,fan2016point,groueix2018atlasnet,kato2018neural,tulsiani2017multi,yan2016perspective}, because creating a large amount of 3D models is very costly, 3D object reconstruction in more diverse categories beyond ShapeNet is difficult. While the majority of methods use 3D shapes as supervision~\cite{choy20163d,fan2016point,groueix2018atlasnet}, several works aim to reduce 3D supervision by using 2D images for training~\cite{kato2018neural,tulsiani2017multi,yan2016perspective}. However, they typically require images with foreground masks of objects, which are still not easy to obtain.

\begin{figure}[t]
    \begin{center}
        \includegraphics[width=1.0\linewidth]{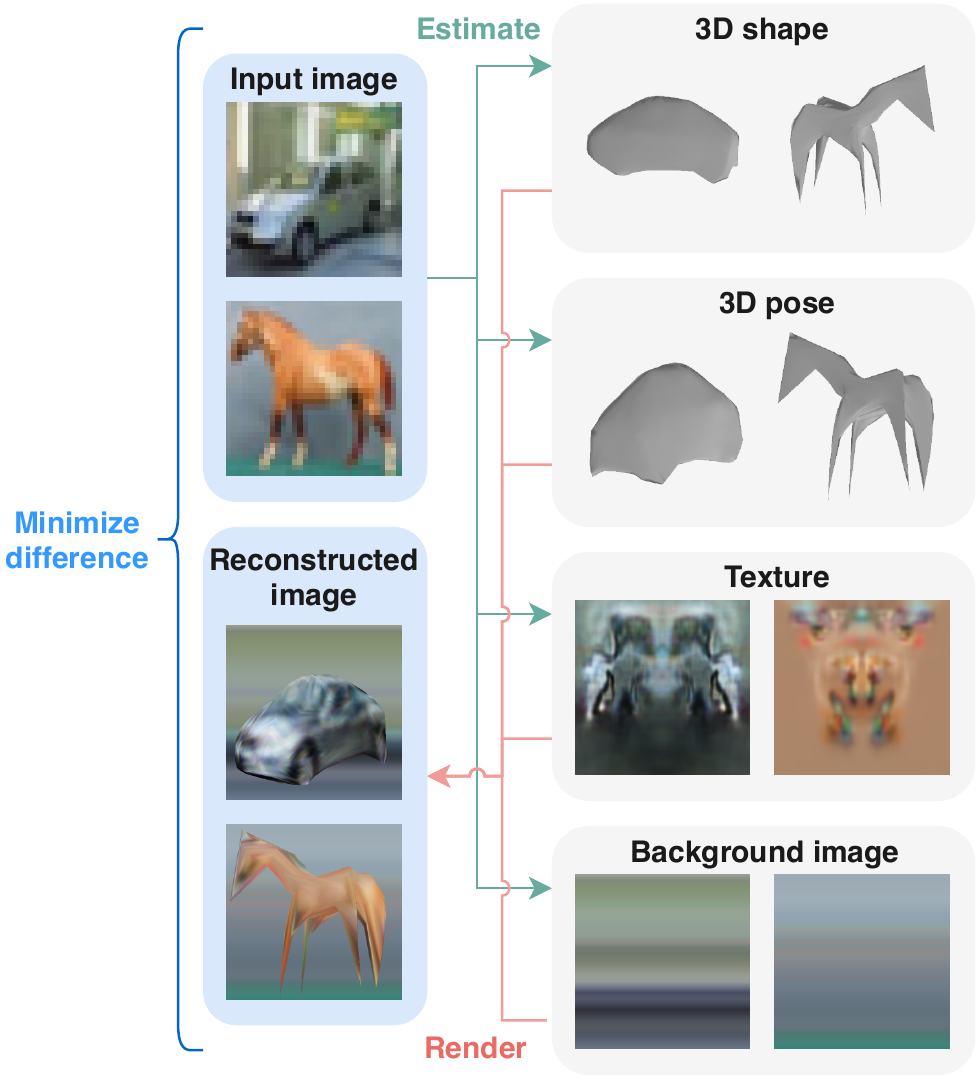}
    \end{center}
    \vspace{-4mm}
    \caption{Given an object image, our proposed model estimates its 3D shape, pose, texture, and background. Only categorized object images are required for training. This figure shows the system architecture and result of our model on CIFAR-10.  }
    \label{fig:top}
    \vspace{-2mm}
\end{figure}

In this work, we demonstrate that 3D shape, pose, and texture can be learned from categorized natural images without supervision. Fig.~\ref{fig:top} shows our result on the test set of CIFAR-10 dataset~\cite{krizhevsky2009learning}. This dataset of categorized natural images has difficulties in several aspects. Ground-truth 3D shapes are not given, there are no multiple views of the same object, foregrounds and backgrounds are not separated, viewpoints are unknown, objects have various shapes and sizes, and they locate and rotate freely. Success in estimating 3D elements on this challenging dataset indicates the possibility to leverage natural images for 3D understanding and realize 3D object reconstruction on diverse object categories.

\begin{figure}[t]
    \begin{center}
        \includegraphics[width=1.0\linewidth]{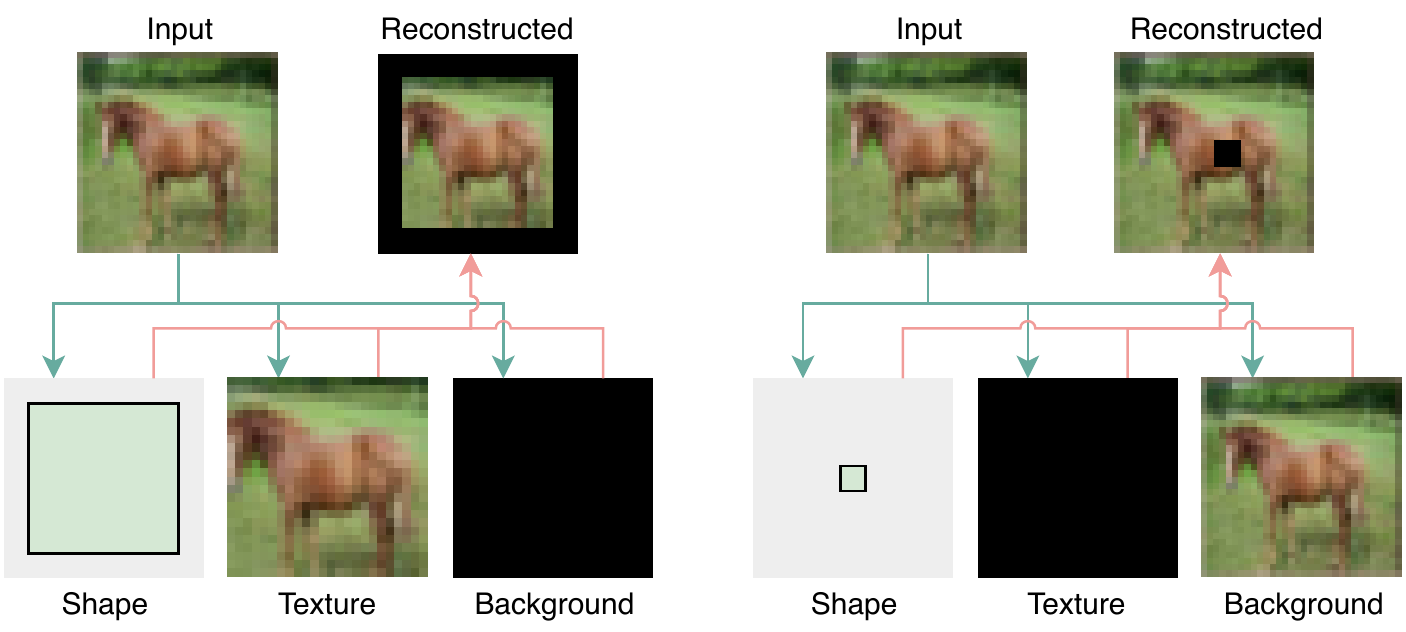}
    \end{center}
    \vspace{-2mm}
    \caption{Examples of trivial and poor solutions. We know these are unrealistic, however, neural networks cannot know it by self-supervision. Therefore, we induct several kinds of structural knowledge of 3D scenes into our model.}
    \label{fig:failures}
    \vspace{-3mm}
\end{figure}

We train this model by comparing input images with reconstructed images. Given an image, 3D shape, pose, texture image, and background image are estimated by neural networks. Then, an image is rendered using these estimated elements. Reconstruction error is computed by extracting and comparing features of the input image and reconstructed image, and neural networks are optimized by minimizing it. Though this framework is quite simple, a way to obtain a meaningful model is not straightforward because this problem has several trivial and poor solutions, as depicted in Fig.~\ref{fig:failures}. One example is to expand an object to cover the whole image and copy the input image into the texture of the object. Another example is to shrink an object under one pixel and copy the input image into the background. Though image reconstruction is almost perfectly in both cases, we know that this reconstruction is unlikely as realistic 3D scenes. The technical key point of this paper is to induct such knowledge about 3D structures into neural networks in the form of training methods, constraints, and regularization. Our main assumptions are (1) all shapes in the same object category are similar, and they can be made by slightly deforming a category-specific base shape, (2) surfaces of objects are smooth, and (3) background images are sufficiently simple. Based on these assumptions, we carefully design constraints and regularization, and separate training steps into category-specific base shape generation and full training given a base shape. While these assumptions are not always correct in real scenes, they are practically useful for training, as we demonstrate in experiments using CIFAR-10 and PASCAL objects~\cite{xiang2014beyond}.

The major contributions can be summarized as follows.
\begin{itemize}
    \setlength\itemsep{0em}
    \vspace{-2mm}
    \item To the best of our knowledge, this is the first study to train single-view reconstruction of 3D shape, pose, and texture by using only 
    categorized natural images.
    \item We demonstrate that self-supervised learning can be achieved by (1) training shapes first and subsequently training poses and textures using the obtained shapes, and (2) introducing regularization and constraints of shapes, textures, and backgrounds.
\end{itemize}

%% file: 2_related_work.tex
\begin{table}[t!]
    \small
    \begin{center}
        \begin{tabular}{lcccccc}
            \toprule
            Supervision  & \cite{rezende2016unsupervised} & \cite{tulsiani2018multi} & \cite{henzler2018escaping} & \cite{zuffi2019three}${}^{\dagger}$ & \cite{nguyen2019hologan}${}^{\ddagger}$ & ours \\
            \hline
            Natural images  &            &            & \checkmark & (\checkmark) & \checkmark & \checkmark \\
            Viewpoint-free  & \checkmark & \checkmark & \checkmark & (\checkmark) & \checkmark & \checkmark \\
            Silhouette-free &            &            &            & (\checkmark) & \checkmark　& \checkmark \\
            \bottomrule
        \end{tabular}
    \end{center}
    \vspace{-1mm}
    \caption{Works that aim supervision reduction in single-view training of single-view 3D object reconstruction. {\it Silhouette-free} includes works that use images without backgrounds. ${}^{\dagger}$Zuffi \etal~\cite{zuffi2019three} leverages simulators. ${}^{\ddagger}$Nguyen-Phuoc \etal~\cite{nguyen2019hologan} cannot represent 3D shapes explicitly.}
    \label{table:relatecd_work}
    \vspace{-2mm}
\end{table}

\section{Related work}
\label{sec:related_work}

There are a vast number of problems and approaches in 3D reconstruction since this is a long-standing topic in computer vision. In this section, we focus on supervision for single-view 3D object reconstruction.

\vspace{-3mm}
\paragraph{3D supervision} When plenty of 3D object models are available, using them as training signals would be the best option because that approach does not suffer from shape ambiguities found in  2D images. This approach has been popularized with the advent of large 3D shape datasets, such as ShapeNet~\cite{chang2015shapenet}. One research direction is how neural networks handle irregular 3D representations, such as high-resolution voxels~\cite{hane2019hierarchical,tatarchenko2017octree}, point clouds~\cite{fan2016point}, meshes~\cite{groueix2018atlasnet,wang2018pixel2mesh} and implicit functions~\cite{chen2019learning,mescheder2019occupancy,park2019deepsdf}. Another path is the generalization of learning algorithms to novel objects~\cite{shin2018pixels,tatarchenko2019single,zhang2018learning}.

\vspace{-3mm}
\paragraph{Multi-view training} Because a 3D shape is understandable from its multiple 2D projections, object silhouettes from multiple viewpoints have been used as alternative training signals. Different from 3D supervision, this view supervision requires a differentiable 3D-to-2D projection module. Therefore, several differentiable projection modules have been developed for voxels~\cite{tulsiani2017multi,yan2016perspective}, meshes~\cite{kato2018neural,liu2019soft}, point clouds~\cite{insafutdinov2018unsupervised}, and implicit functions~\cite{liu2019learning,sitzmann2019scene}. Though annotation cost is lower than 3D supervision, multiple views are still costly because collecting them requires a specialized 3D capture system.

\vspace{-3mm}
\paragraph{Single-view training} Training using image collections would be the lowest cost choice. However, since it is not an easy task, additional supervision is typically required. Kar \etal~\cite{kar2015category} demonstrated that 3D shapes and viewpoints can be recovered from natural images when silhouettes and keypoints of objects are available. Kanazawa \etal~\cite{kanazawa2018learning} translated this framework into neural networks and incorporated texture prediction in addition. Tulsiani \etal~\cite{tulsiani2017multi} applied their multi-view training method that uses silhouette and viewpoint supervision onto a single-view dataset. Later, they relaxed this dataset requirement by integrating pose prediction~\cite{tulsiani2018multi}. Kato and Harada~\cite{kato2019learning} demonstrated that a similar approach tends to result in unrealistic-looking shapes and improved it by adversarial training. Rezende \etal~\cite{rezende2016unsupervised} trained 3D structure from images, however, their dataset is composed of simple primitives without backgrounds. Henzler \etal~\cite{henzler2018escaping} trained 3D object reconstruction from natural images with silhouette annotations. Nguyen-Phuoc \etal~\cite{nguyen2019hologan} learned implicit neural 3D representation from natural images. Zuffi \etal~\cite{zuffi2019three} leveraged simulators to learn rare 3D objects. Contrary to these works, our method can learn explicit 3D structures from natural images, and it does not require any supervision except for categorized object images. Table~\ref{table:relatecd_work} shows a summary.

%% file: 3_method.tex
\begin{figure}[t]
    \begin{center}
        \includegraphics[width=1.0\linewidth]{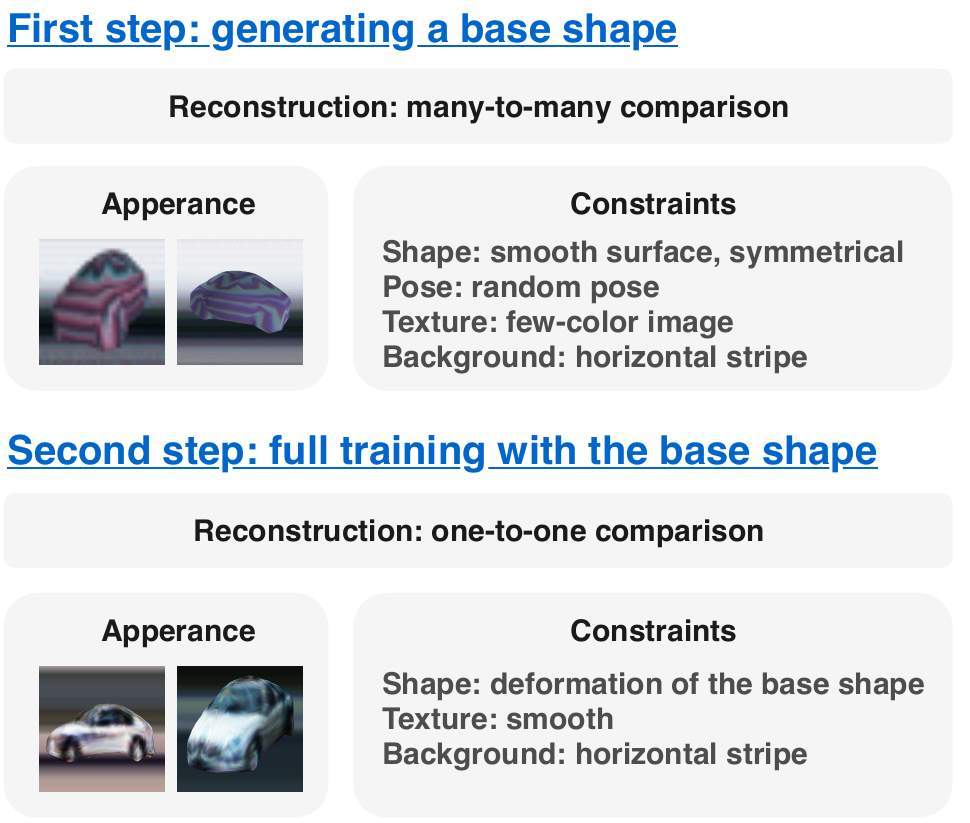}
    \end{center}
    \vspace{-2mm}
    \caption{Training steps and constraints in our proposed method. }
    \label{fig:system5}
    \vspace{-2mm}
\end{figure}

\section{Method}
\label{sec:method}

Our method trains single-view reconstruction of 3D shape, pose, texture, and background with self-supervision as shown in Fig.~\ref{fig:top} while avoiding unrealistic solutions like those shown in Fig.~\ref{fig:failures}. One difficulty is training all elements at the same time because neural networks easily fall in the easiest solution of copying an input image into pixel arrays (textures or backgrounds). Therefore, we propose a two-stage training method that focuses on shapes first. Fig.~\ref{fig:system5} illustrates the overview of our proposed approach. In the first step, a category-specific 3D base shape is generated by maximizing the similarity between images in a dataset and images of the shape. We use randomly sampled viewpoints and strongly limited textures. In the second step, the whole model is trained limiting generated shapes to deformations of the obtained base shape. Another difficulty is that shapes and backgrounds sometimes become excessively complicated to mistakenly reconstruct textures on object surfaces. To suppress it, we propose using strong regularization and constraints on object surfaces and background images. We use a mesh as a 3D representation to introduce constraints and regularization on texture and surfaces.

\subsection{Learning category-specific base shape}
\label{sec:method_base_shape}

\begin{figure}[t]
    \begin{center}
    \includegraphics[width=1.0\linewidth]{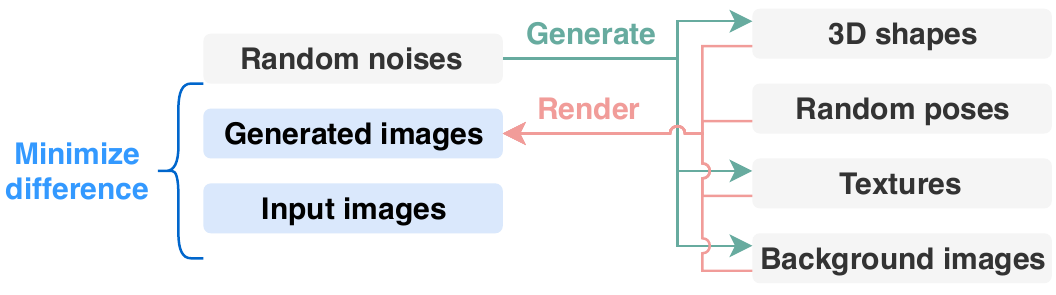}
    \end{center}
    \vspace{-2mm}
      \caption{Our proposed model for learning category-specific base shapes. To focus on learning shapes, viewpoints are randomly sampled from a fixed distribution, and the representation capacity of textures is limited.}
    \label{fig:system2}
    \vspace{-2mm}
\end{figure}

In training a category-specific base 3D shape, we strongly limit the representation capacity of textures in order to focus on shapes. Because this limitation makes it impossible to reconstruct images with results close to the input images, adoption of an auto-encoder architecture in Fig.~\ref{fig:top} is infeasible. Instead, as shown in Fig.~\ref{fig:system2}, we propose a model that generates a shape, texture, and background from random noise by minimizing the difference between the set of rendered images and the set of images in a dataset. In the following sections, we explain each component along with the additional constraints and regularization needed to obtain a meaningful shape. 

\subsubsection{Shape generation}
\label{sec:method_pt_shape}
We generate a shape by deforming vertices of a pre-defined sphere, as was done in several existing works~\cite{kanazawa2018learning,kato2018neural,wang2018pixel2mesh}. Additionally, we manually set an initial dimension of the sphere in each category. With a pre-defined shape of $N_v$ vertices, $N_v \times 3$ variables are generated by a neural network and added to vertex coordinates in 3D space. Then, the generated shape is scaled to fit into a unit cube. In addition, we employ the following constraints and regularization.

\vspace{-3mm}
\paragraph{Smoothness of objects}
Because the representation capacity of textures is limited, the generated shapes try to be very complicated in order to represent edges in images. However, most of the edges in natural images are actually caused by textures or backgrounds, not by shapes. Therefore, we assume that the surfaces of objects are smooth and regularize curvature of them, which is a common approach in modeling object surfaces~\cite{barron2014shape,kanazawa2018learning}. Specifically, we minimize graph Laplacian of a mesh, which represents approximated mean curvature at each vertex~\cite{taubin1995signal}, and angles between two neighboring triangle polygons, which implies smoothness at each edge. We denote this loss term as $\mathcal{L}_s$. More details are in the appendix.

\vspace{-3mm}
\paragraph{Symmetry of objects} Though object shapes in natural images are not always symmetrical (e.g. horses), category-specific base shapes are often symmetrical (e.g. the average shape of horses). Therefore, we constrain generated shapes to be symmetrical. 

\subsubsection{Texture generation}
\label{sec_method_texture_pt}

We assume that UV-mapping of a texture image and surface is pre-defined and fixed during training. To generate a texture image, we employ DCGAN~\cite{radford2015unsupervised}-like architecture with residual connections~\cite{he2016deep}. 

\vspace{-3mm}
\paragraph{Simplicity of object textures} To reduce the representation capacity of texture images, we propose using a single color or only a few colors when making a texture image. A similar assumption is used for reflectance maps in intrinsic image decomposition~\cite{barron2014shape}. Specifically, instead of generating a three-channel RGB image, we generate a $N_{\text{c}}$-channel image by a neural network and normalize each pixel, so that the sum of the channel values is one. Additionally, a color palette of $N_{\text{c}}$ colors is generated by another neural network. Then, an RGB image is generated by mixing the $N_{\text{c}}$ colors according to the $N_{\text{c}}$-channel image. Though the representation capability of this reparameterization is the same as the original network when $N_{\text{c}} \geq 3$, it generates few-color images in practice.

\subsubsection{Pose generation}
\label{sec:method_pt_pose}
To represent the 6DoF pose of an object, we assume that a camera is always directed to the center of the object, the upward direction of the camera is always $(x, y, z) = (0, 1, 0)$, the distance between the object and the camera is fixed, and only azimuth and elevation of viewpoints can be changed. During training, viewpoints are sampled randomly from a manually-designed viewpoint distribution. 

\subsubsection{Background generation}

To prevent the solution shown in the right of Fig.~\ref{fig:failures}, we need to limit the representation capability of background images. Therefore, we introduce the following constraint.

\vspace{-3mm}
\paragraph{No vertical lines in backgrounds} The representation capacity of a background image must be high enough to express the structure of a scene, however, it must not be too high in order not to represent foreground objects. To achieve this, we constrain background images to be horizontal stripes without vertical lines. Images with this constraint can express the rough scene structure, such as the sky and grasses, however, they cannot express objects.

\subsubsection{Rendering}
We render an image using a generated shape, pose, texture, and background with random directional lighting and smooth shading. Using light is essential to express shapes with limited few-color textures. We use a differentiable renderer developed by Kato \etal\cite{kato2018neural} to back-propagate the gradient from the loss function into generators.

\subsubsection{Comparison between real and generated images} 
\label{sec:method_pt_reconstruction}
Since one-to-one comparison of real and generated images is impossible, we match distributions of them similar to generative adversarial networks (GANs)~\cite{goodfellow2014generative}. However, adversarial training used in GANs does not work well for our problem because the limited representation capacity of the image generator makes the minimax game too advantageous for discriminators. Instead, we use feature matching~\cite{salimans2016improved} and the Chamfer distance of real and generated image minibatches to compute a reconstruction loss $\mathcal{L}_{\text{rec}}$. More details are in the appendix.

\subsubsection{Summary and post-processing} 
In this step, to obtain a category-specific base shape, a shape generator, a texture generator, and a background generator are trained by minimizing the sum of reconstruction loss $\mathcal{L}_{\text{rec}}$ and smoothing loss $\mathcal{L}_{\text{s}}$ under the constraints of shape symmetricity, and texture and background simplicity. Though the input is random noise, generated shapes converge to a single shape after training, which is similar to mode collapse in GANs.

For texture mapping, a smaller variance of polygon sizes is better. To accomplish this, we use silhouettes of the obtained mesh to generate another mesh by minimizing several factors: the difference of the silhouettes, the variance of sizes of the triangle polygons, and the whole area of the surfaces. This post-processing significantly reduces the variance of polygon sizes while maintaining the whole shape.

\subsection{Full training with base shape}

In the second step, we train the pipeline in Fig.~\ref{fig:top} while limiting the generated shapes to deformations of a category-specific base shape. We use encoder-decoder architecture for shape, pose, texture, and background prediction. We use a texture generator without the few-color constraint because providing base shapes prevents results like those found in the left of Fig.~\ref{fig:failures}. In addition, we regularize the total variation~\cite{rudin1992nonlinear} of textures to reduce noise. We also use the same background generator as we did for the previous step so as to prevent results like those found in the right of Fig.~\ref{fig:failures}. We render images without using directional lighting because textures are able to represent shadings in this step.

\subsubsection{Shape prediction} 
Instead of predicting a mesh directly, we predict shape deformations using free-form deformation~\cite{sederberg1986free} similar to several other object reconstruction works~\cite{kurenkov2018deformnet,yumer2016learning}. We use a spatial grid of $4 \times 4 \times 4$ vertices, and regress the difference between the original grid and a deformed grid using a neural network. In addition, we use another network to regress the relative height, width, and length of shapes. After deformation, the size of the predicted shape is scaled to fit a unit cube.

\vspace{-3mm}
\paragraph{Exploring best shape} The variation between generated shapes tends to be very small because exploring various shapes using only a differentiable renderer and gradient descent is difficult due to local minima. To overcome this problem, we explore and record the best shape for each input image at each training iteration. Specifically, we render images using an estimated shape, a recorded best shape, a slightly perturbed the best shape, and random shapes. Then, we compute reconstruction loss to find the best one and record it.

\subsubsection{Pose prediction} 
In this step, we parameterize the 6DoF object/camera pose by azimuth and elevation as with Section~\ref{sec:method_pt_pose}, in-plane rotation of an object, center point of an object in 2D image coordinates, and scale of an object. We train a decoder that outputs these six parameters. We adopt multiple regressor approach used in~\cite{insafutdinov2018unsupervised}.

\vspace{-3mm}
\paragraph{Exploring best pose} Similarly to shape prediction, we also need to actively explore the best poses. At each training iteration, we explore and record the best pose for each input image by rendering images using estimated, recorded, random, and perturbed poses.

\subsubsection{Training}

In addition to the components described above, we employ view prior learning (VPL)~\cite{kato2019learning} to reduce overfitting to the observed views. Summarily, a loss function is composed of the following four terms. (1) Reconstruction loss. Reconstructed images using the best shapes, estimated textures, the best poses, and estimated backgrounds are compared with input images. In addition, feature matching is also used. (2) Mean absolute error between estimated shapes/poses and the best shapes/poses that are recorded during training. (3) Total variation of estimated texture images for denoising. (4) VPL loss. To facilitate an early phase of training, at the $i$-th iteration, training samples are randomly selected from first to $i$-th data in the dataset. This makes the model see the same sample frequently in an early stage, which simplifies finding the best poses and makes the estimated poses diverse.

\subsubsection{Photometric per-instance fine-tuning}

In inference, similar to~\cite{zuffi2019three}, we slightly adjust predictions by optimizing the outputs of encoders to minimize the reconstruction loss. This is possible because we do not need silhouette or viewpoint annotations to compute the loss. We successively optimize the outputs of the background encoder, pose encoder, and shape decoder.

%% file: 4_experiments.tex
\begin{figure}[t]
    \begin{center}
        \parbox{70mm}{(a) W/ all constraints} \vspace{1mm} \\
        \includegraphics[width=15mm]{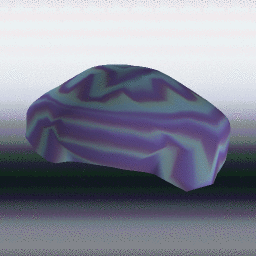}
        \includegraphics[width=15mm]{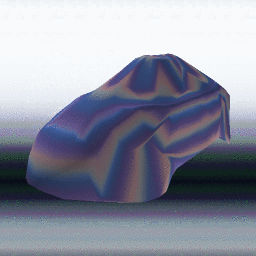} 
        \includegraphics[width=15mm]{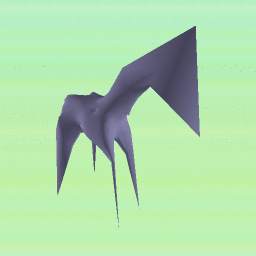}
        \includegraphics[width=15mm]{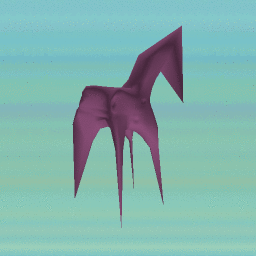} \\
        \parbox{70mm}{(b) W/o shape smoothness } \vspace{1mm} \\
        \includegraphics[width=15mm]{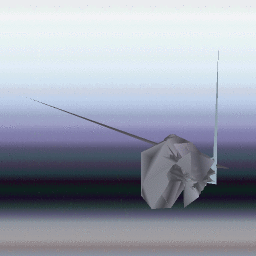}
        \includegraphics[width=15mm]{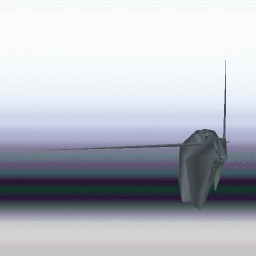} 
        \includegraphics[width=15mm]{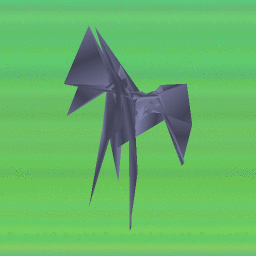}
        \includegraphics[width=15mm]{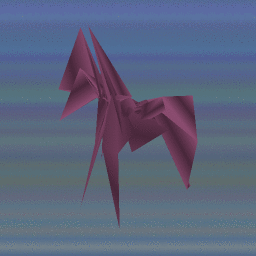} \\
        \parbox{70mm}{(c) W/o shape symmetricity} \vspace{1mm} \\
        \includegraphics[width=15mm]{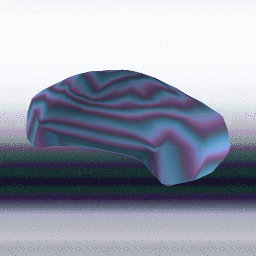}
        \includegraphics[width=15mm]{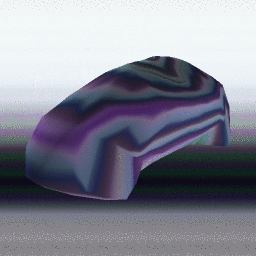} 
        \includegraphics[width=15mm]{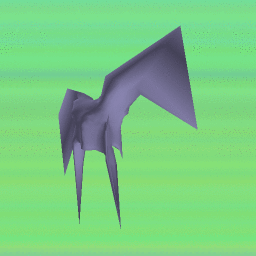}
        \includegraphics[width=15mm]{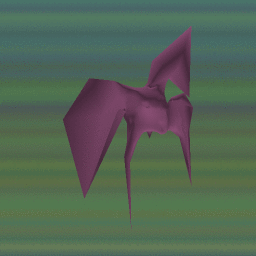} \\
        \parbox{70mm}{(d) W/o texture simplicity } \vspace{1mm} \\
        \includegraphics[width=15mm]{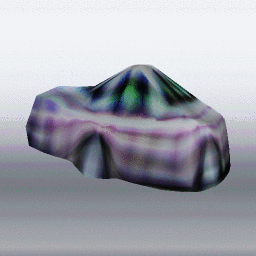}
        \includegraphics[width=15mm]{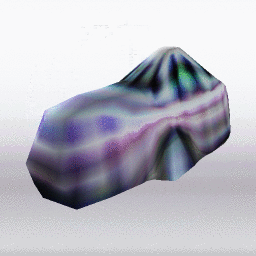} 
        \includegraphics[width=15mm]{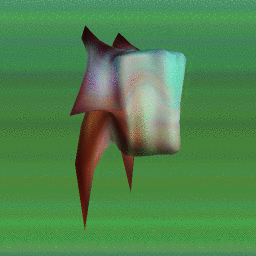}
        \includegraphics[width=15mm]{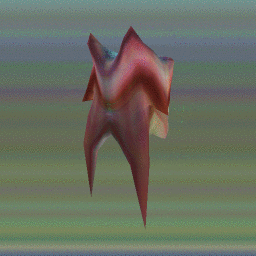} \\
        \parbox{70mm}{(e) W/o background simplicity} \vspace{1mm} \\
        \includegraphics[width=15mm]{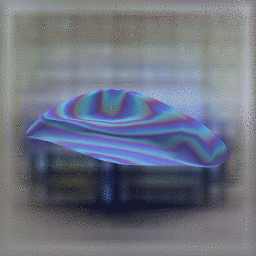}
        \includegraphics[width=15mm]{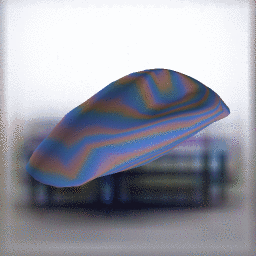} 
        \includegraphics[width=15mm]{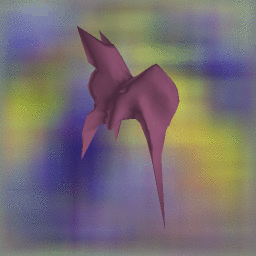}
        \includegraphics[width=15mm]{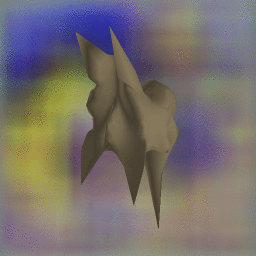} \\
    \end{center}
    \caption{Generated category-specific base shapes on CIFAR-10 dataset. These images are rendered in $256 \times 256$ resolution with upsampled background images. (a) shows a result of our proposed method, and (b--e) are ablation studies that clarify contributions of introduced constraints and regularization.}
    \label{fig:exp_pt}
\end{figure}

\begin{figure*}[t]
    \begin{center}
        \parbox{45mm}{\raisebox{14mm}{(a) Input images}}
        \includegraphics[width=15mm]{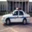}
        \includegraphics[width=15mm]{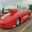}
        \includegraphics[width=15mm]{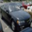}
        \includegraphics[width=15mm]{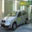}
        \includegraphics[width=15mm]{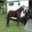}
        \includegraphics[width=15mm]{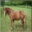}
        \includegraphics[width=15mm]{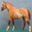}
        \includegraphics[width=15mm]{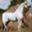} \vspace{-7mm} \\
        \parbox{45mm}{\raisebox{14mm}{\parbox{45mm}{(b) Reconstructed images \\ w/o photometric fine-tuning}}}
        \includegraphics[width=15mm]{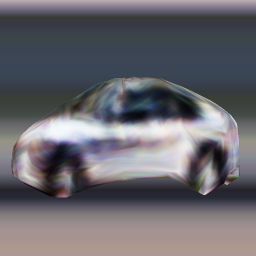}
        \includegraphics[width=15mm]{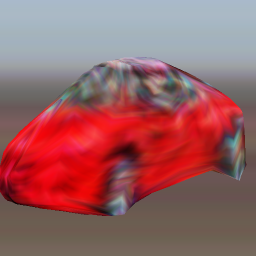}
        \includegraphics[width=15mm]{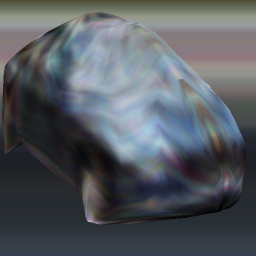}
        \includegraphics[width=15mm]{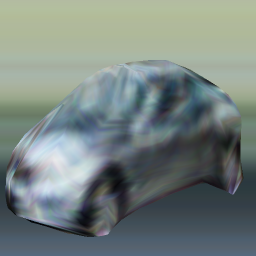}
        \includegraphics[width=15mm]{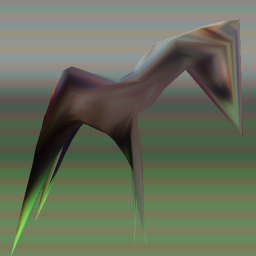}
        \includegraphics[width=15mm]{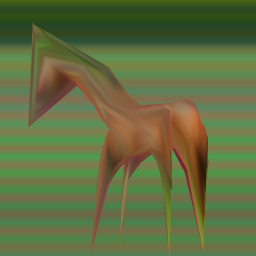}
        \includegraphics[width=15mm]{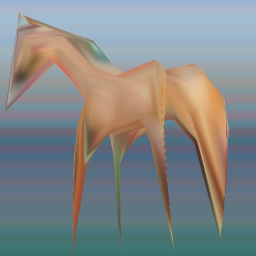}
        \includegraphics[width=15mm]{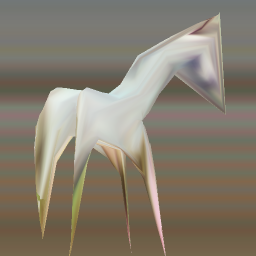} \vspace{-8.0mm} \\
        \parbox{45mm}{\raisebox{14mm}{\parbox{45mm}{(c) Reconstructed images \\ w/ photometric fine-tuning}}}
        \includegraphics[width=15mm]{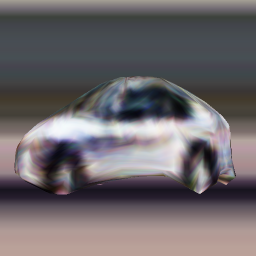}
        \includegraphics[width=15mm]{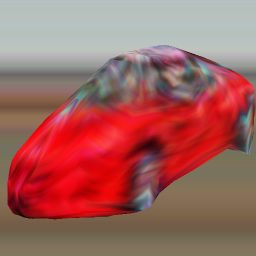}
        \includegraphics[width=15mm]{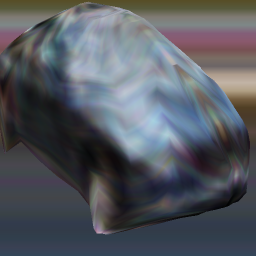}
        \includegraphics[width=15mm]{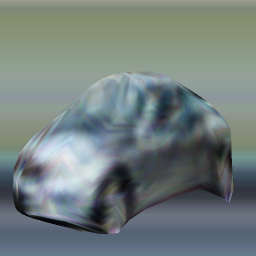}
        \includegraphics[width=15mm]{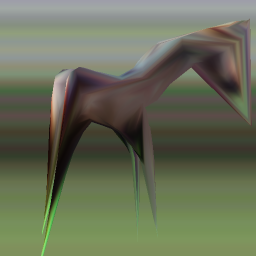}
        \includegraphics[width=15mm]{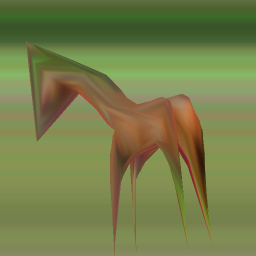}
        \includegraphics[width=15mm]{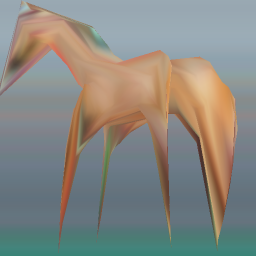}
        \includegraphics[width=15mm]{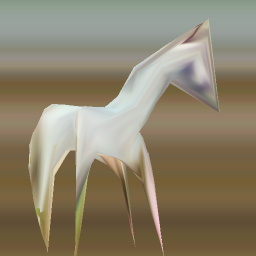} \vspace{-8.0mm} \\
        \parbox{45mm}{\raisebox{14mm}{\parbox{45mm}{(d) 3D models of (c) \\ from another viewpoint}}}
        \includegraphics[width=15mm]{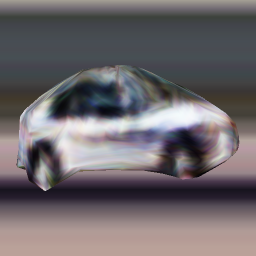}
        \includegraphics[width=15mm]{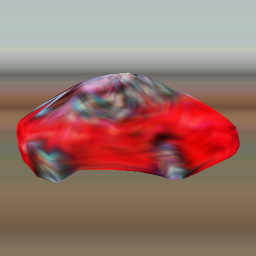}
        \includegraphics[width=15mm]{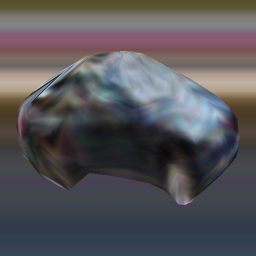}
        \includegraphics[width=15mm]{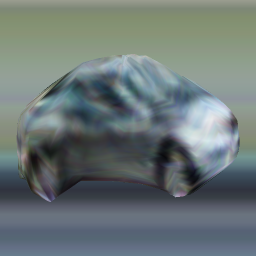}
        \includegraphics[width=15mm]{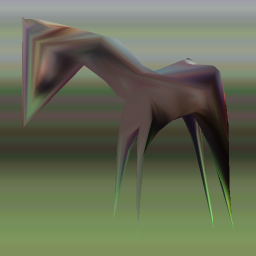}
        \includegraphics[width=15mm]{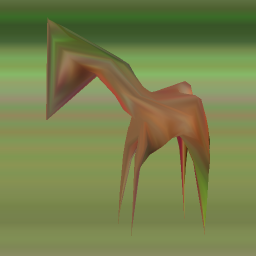}
        \includegraphics[width=15mm]{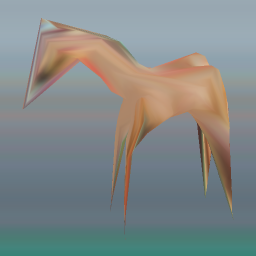}
        \includegraphics[width=15mm]{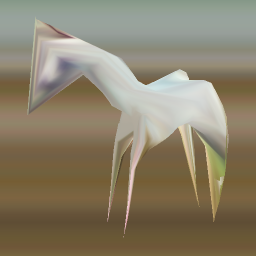} \vspace{-8.0mm} \\
    \end{center}
    \caption{Representative results of 3D shape, pose, texture, and background estimation on CIFAR-10 test set. To understand the shapes and textures better, images are rendered at a higher resolution with upsampled backgrounds. Since the input images (a) are explicitly disentangled into 3D object elements, objects can be rendered from another viewpoint (d). Randomly selected results are in the appendix. }
    \label{fig:exp_ft}
\end{figure*}

\begin{figure}[t]
    \begin{center}
        \parbox{45mm}{\raisebox{14mm}{\parbox{45mm}{(a) Input images}}}
        \includegraphics[width=15mm]{./images/cifar10_ft_car/0006_in.png}
        \includegraphics[width=15mm]{./images/cifar10_ft_horse/0002_in.png}
        \vspace{-7mm} \\
        \parbox{45mm}{\raisebox{14mm}{\parbox{45mm}{(b) Reconstructed images \\ w/ constraints}}}
        \includegraphics[width=15mm]{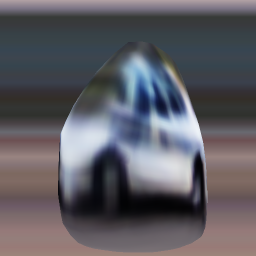}
        \includegraphics[width=15mm]{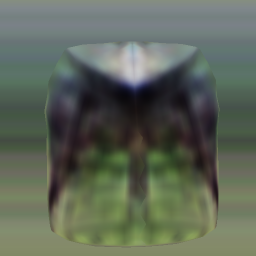}
        \vspace{-8mm} \\
        \parbox{45mm}{\raisebox{14mm}{\parbox{45mm}{(c) 3D models of (b) \\ from another viewpoint}}}
        \includegraphics[width=15mm]{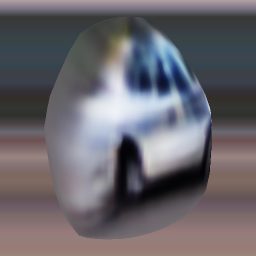}
        \includegraphics[width=15mm]{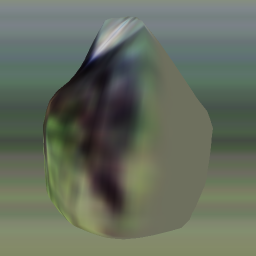}
        \vspace{-8mm} \\
        \parbox{45mm}{\raisebox{14mm}{\parbox{45mm}{(d) Reconstructed images \\ w/o constraints}}}
        \includegraphics[width=15mm]{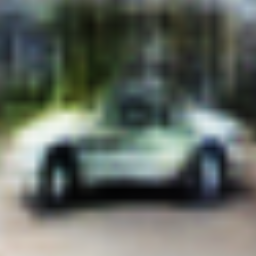}
        \includegraphics[width=15mm]{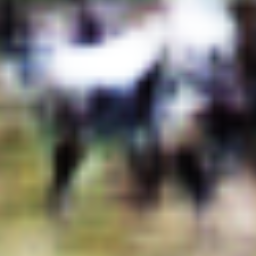}
        \vspace{-8mm} \\
        \parbox{45mm}{\raisebox{14mm}{\parbox{45mm}{(e) 3D models of (d) \\ from another viewpoint}}}
        \includegraphics[width=15mm]{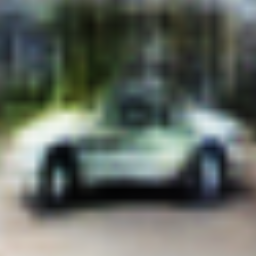}
        \includegraphics[width=15mm]{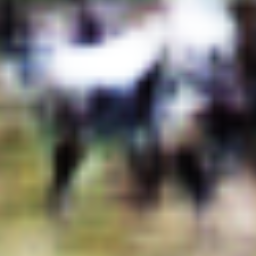}
        \vspace{-8mm} \\
    \end{center}
    \caption{Training without our proposed two-stage training. (b--c) and (d--e) correspond to the left and right of Fig.~\ref{fig:failures} respectively. These results confirm the importance of training shapes explicitly. }
    \label{fig:exp_single_stage}
\end{figure}

\begin{figure*}[t]
    \begin{center}
        \parbox{45mm}{\raisebox{14mm}{(a) Ours (base shape)}}
        \includegraphics[width=15mm]{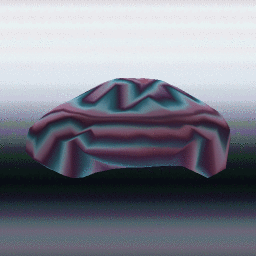}
        \includegraphics[width=15mm]{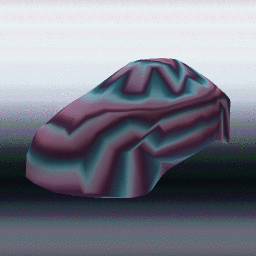}
        \includegraphics[width=15mm]{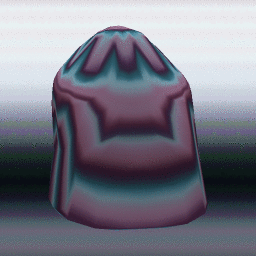}
        \includegraphics[width=15mm]{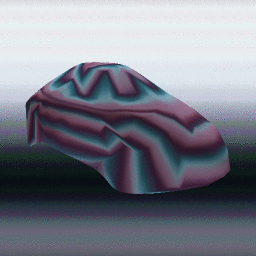}
        \includegraphics[width=15mm]{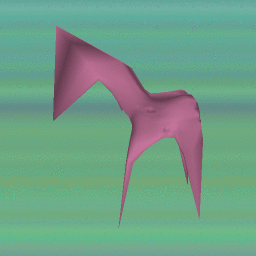}
        \includegraphics[width=15mm]{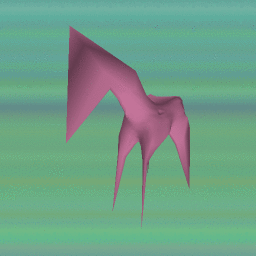}
        \includegraphics[width=15mm]{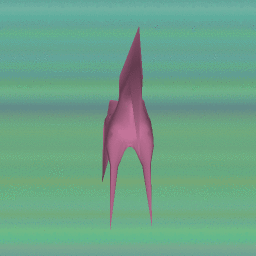}
        \includegraphics[width=15mm]{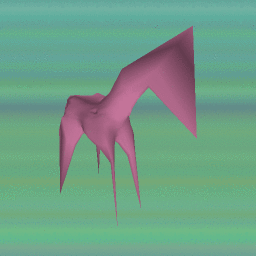} \vspace{-7.0mm} \\
        \parbox{45mm}{\raisebox{14mm}{(b) Ours (full model)}}
        \includegraphics[width=15mm]{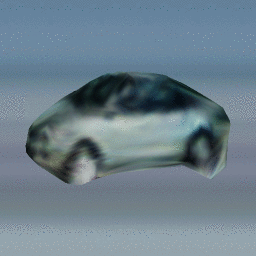}
        \includegraphics[width=15mm]{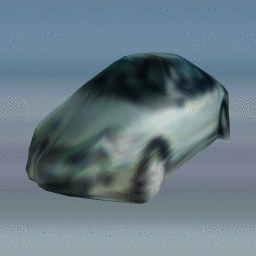}
        \includegraphics[width=15mm]{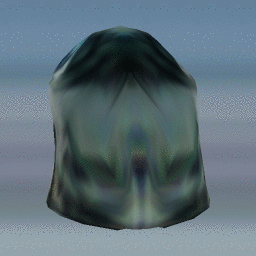}
        \includegraphics[width=15mm]{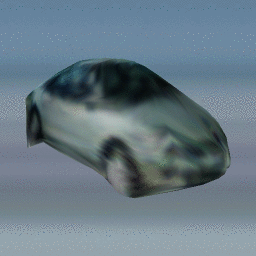}
        \includegraphics[width=15mm]{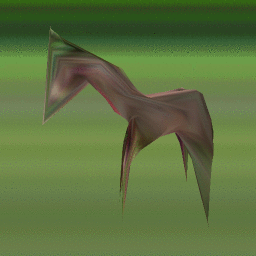}
        \includegraphics[width=15mm]{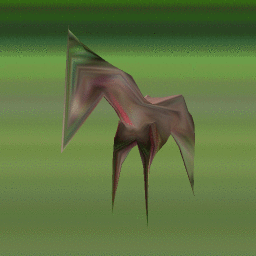}
        \includegraphics[width=15mm]{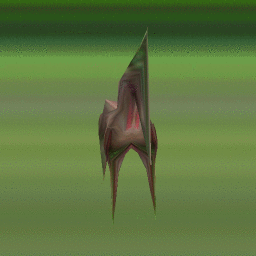}
        \includegraphics[width=15mm]{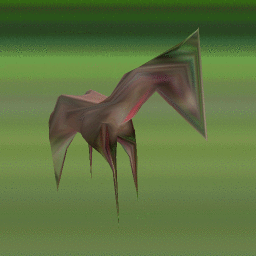} \vspace{-7.0mm} \\
        \parbox{45mm}{\raisebox{14mm}{(c) HoloGAN on CIFAR-10}}
        \includegraphics[width=15mm]{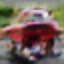}
        \includegraphics[width=15mm]{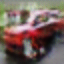}
        \includegraphics[width=15mm]{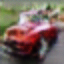}
        \includegraphics[width=15mm]{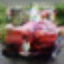}
        \includegraphics[width=15mm]{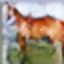}
        \includegraphics[width=15mm]{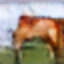}
        \includegraphics[width=15mm]{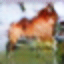}
        \includegraphics[width=15mm]{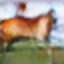} \vspace{-7.0mm} \\
    \end{center}
    \caption{Comparison of ours with HoloGAN on CIFAR-10. Images are rendered at 50 degree intervals.}
    \label{fig:hologan_cifar10}
\end{figure*}

\section{Experiments}

\subsection{CIFAR-10}

We mainly tested our method on the CIFAR-10~\cite{krizhevsky2009learning} dataset because it is composed of natural images and contains thousands of images per object category. Among ten object categories, we focused on {\it car} and {\it horse} classes because {\it car} is an artificial and rigid object and one of the most commonly used categories on the synthetic ShapeNet dataset~\cite{chang2015shapenet} and {\it horse} is a deformable natural object not contained in ShapeNet. For feature extraction, we trained WRN-16-4~\cite{zagoruyko2016wide} on the CIFAR-10 training set. We used three layers right before sub-sampling as feature maps.

\subsubsection{Base shape learning}

First, we evaluate the first step described in Section~\ref{sec:method_base_shape}. We set the number of colors of {\it car} texture to four, and that of {\it horse} to one. We trained $10$ models for each category using different random seed and selected the best-looking one. Fig.~\ref{fig:exp_pt} (a) shows generated base shapes by our proposed method using different random noise. The generated shapes, textures, and backgrounds look plausible. Particularly, the {\it horse} correctly has four legs, and the {\it car} has four tires on texture. The background image of {\it car} represents the sky as a bright region and roads as dark, and the background of {\it horse} shows grasses. This result indicates that the generators work properly.

Fig.~\ref{fig:exp_pt} (b--e) shows ablation study. When the regularization of shape smoothness is removed, thin lines are generated to represent edges, which results in unrealistic shapes (b). The shape symmetricity constraint seems unimportant for {\it car}, but it helps to generate legs on {\it horse} regularly. (c). Even when the texture simplicity constraint is removed, the texture does not represent the whole scene as in the left of Fig.~\ref{fig:failures} because of constraints on shapes. However, the texture of {\it horse} contains the colors of horses and grasses, that results in the incorrect shape (d). When the background simplicity constraint is not used, the background generator tries to represent shapes, especially in {\it horse}  (e). These results indicate introducing our knowledge about 3D scenes into a model is essential in self-supervised shape learning, and all of the constraints and regularization used are indispensable.

\subsubsection{Full training using base shapes}

Secondly, we evaluate the second step using the base shapes obtained in the previous step. Fig.~\ref{fig:exp_ft} shows representative results on the test set. Reconstructed images demonstrate that the estimators trained by our method are able to reconstruct images that look similar to input images (a--b). Estimated shapes, poses, and backgrounds can be further improved by simple gradient descent and photometric reconstruction loss (c). Rendered images from other viewpoints show that these objects have correct 3D shapes, which are slightly different among different input images (d).

\subsubsection{Effectiveness of two-stage training}
\label{sec:exp_two_stages}
One of the most important techniques of our method is to separate training into two stages. To validate its effectiveness, we trained our models in a single stage. Fig.~\ref{fig:exp_single_stage} shows the reconstruction results by the learned models. When our proposed constraints, such as surface smoothness and background simplicity, are used, the textures represent edges of shapes, which results in incorrect shapes (a--b). When these constraints are not used, because the background estimator copies input images, the reconstructed images look the same from any viewpoint (c--d). Apparently, neither model understands these 3D scenes correctly. These results correspond to the left and right of Fig.~\ref{fig:failures} respectively.

\subsubsection{Comparison with existing works}
To the best of our knowledge, this is the first work to learn single-image reconstruction of 3D shape, pose, and texture from natural image collections without supervision. Therefore, we cannot conduct fair comparison between our work and existing works. One related approach would be structure-from-motion because it can recover an object shape from a photo collection. Therefore, we tested COLMAP~\cite{schoenberger2016sfm,schoenberger2016mvs} on CIFAR-10, however, it failed to reconstruct a shape because it cannot find initial corresponding image pairs. Though slightly different from 3D reconstruction, HoloGAN~\cite{nguyen2019hologan} learns a generative model of images with implicit, but manipulable, 3D representation from natural images. Fig.~\ref{fig:hologan_cifar10} shows comparison between images of our base shapes and images by HoloGAN trained on CIFAR-10. While the shapes produced by our method are consistent from multiple viewpoints, the shapes produced by HoloGAN are not. This result implies the effectiveness of having explicit 3D representations.

\begin{figure*}[t]
    \begin{center}
        \includegraphics[width=15mm]{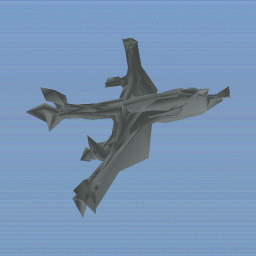}
        \includegraphics[width=15mm]{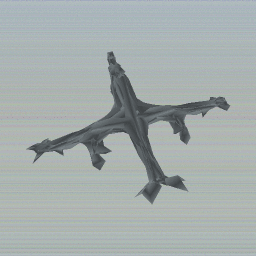}
        \includegraphics[width=15mm]{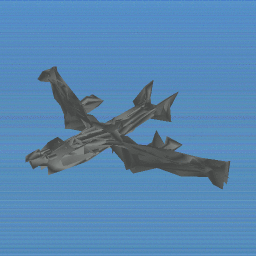}
        \includegraphics[width=15mm]{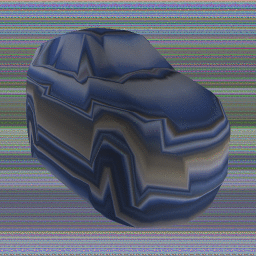}
        \includegraphics[width=15mm]{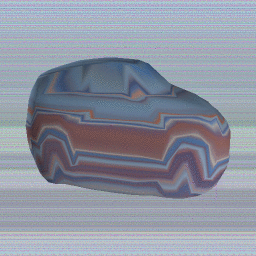}
        \includegraphics[width=15mm]{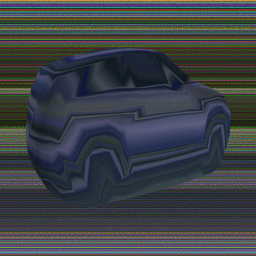}
        \includegraphics[width=15mm]{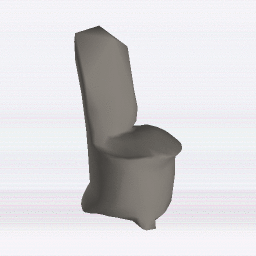}
        \includegraphics[width=15mm]{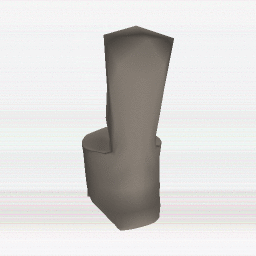}
        \includegraphics[width=15mm]{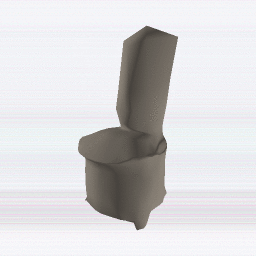}
    \end{center}
    \caption{Generated base shapes of {\it aeroplane}, {\it car}, and {\it chair} on PASCAL dataset.}
    \label{fig:pt_pascal}
\end{figure*}

\begin{figure*}[t!]
    \begin{center}
        \parbox{30mm}{\raisebox{14mm}{(a) Input}}
        \includegraphics[width=15mm]{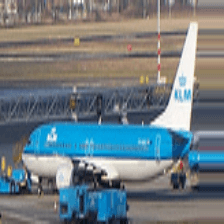}
        \includegraphics[width=15mm]{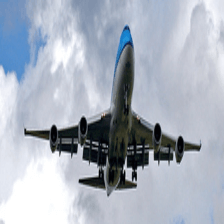}
        \includegraphics[width=15mm]{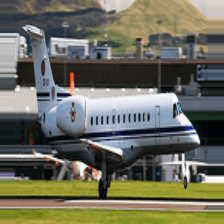}
        \includegraphics[width=15mm]{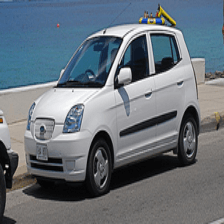}
        \includegraphics[width=15mm]{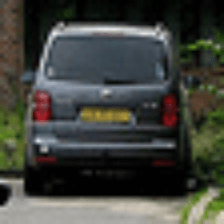}
        \includegraphics[width=15mm]{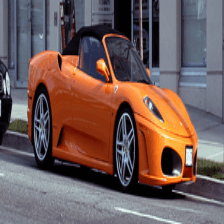}
        \includegraphics[width=15mm]{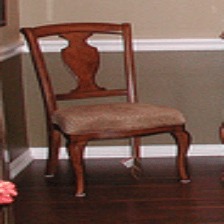}
        \includegraphics[width=15mm]{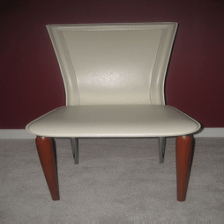}
        \includegraphics[width=15mm]{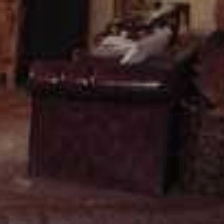}
        \vspace{-7mm} \\
        \parbox{30mm}{\raisebox{14mm}{\parbox{30mm}{(b) Reconstructed}}}
        \includegraphics[width=15mm]{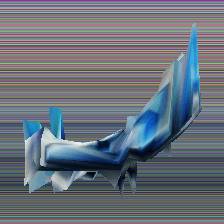}
        \includegraphics[width=15mm]{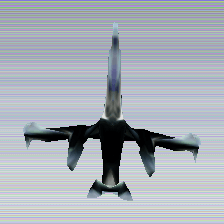}
        \includegraphics[width=15mm]{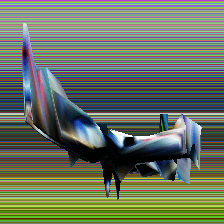}
        \includegraphics[width=15mm]{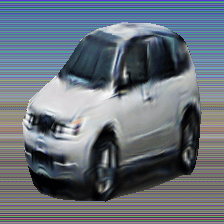}
        \includegraphics[width=15mm]{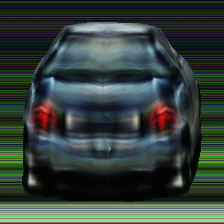}
        \includegraphics[width=15mm]{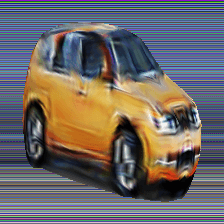}
        \includegraphics[width=15mm]{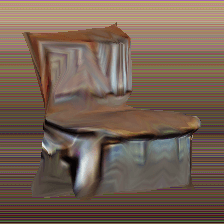}
        \includegraphics[width=15mm]{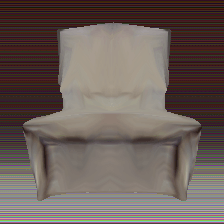}
        \includegraphics[width=15mm]{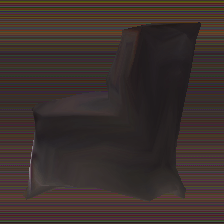}
        \vspace{-7.0mm} \\
        \parbox{30mm}{\raisebox{14mm}{\parbox{30mm}{(c) Reconstructed \\ (another viewpoint)}}}
        \includegraphics[width=15mm]{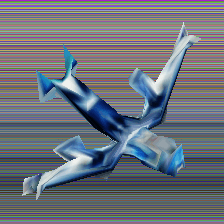}
        \includegraphics[width=15mm]{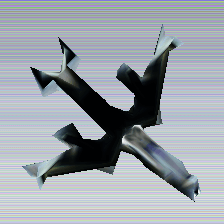}
        \includegraphics[width=15mm]{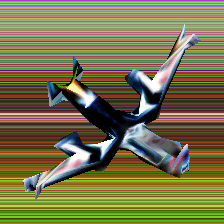}
        \includegraphics[width=15mm]{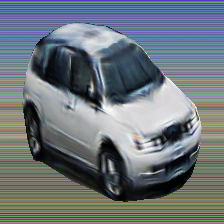}
        \includegraphics[width=15mm]{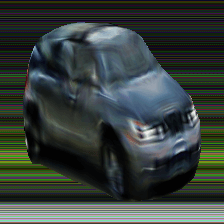}
        \includegraphics[width=15mm]{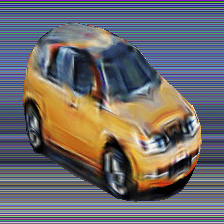}
        \includegraphics[width=15mm]{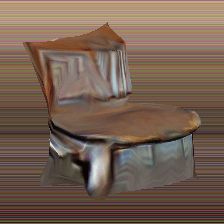}
        \includegraphics[width=15mm]{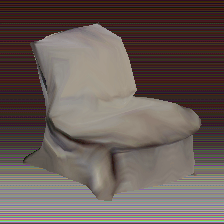}
        \includegraphics[width=15mm]{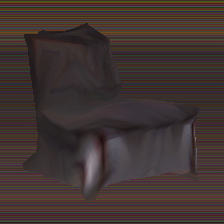}
        \vspace{-8.0mm} \\
    \end{center}
    \caption{Representative results of 3D shape, pose, texture, and background estimation on PASCAL validation set. Randomly sampled results are in the appendix. }
    \label{fig:exp_ft_pascal}
\end{figure*}

\begin{table}[t!]
    \small
    \begin{center}
        \begin{tabular}{l@{\hspace{0.5cm}}l@{\hspace{0.5cm}}ccc}
            \toprule
            Method & Set & {\it airplane} & {\it car} & {\it chair} \\
            \hline
            \multicolumn{5}{l}{With pose and keypoint supervision} \\
            K\&V~\cite{tulsiani2015viewpoints} & validation    & $.81$ & $.90$ & $.80$  \\
            \hline
            \multicolumn{5}{l}{Self-supervised} \\
            Ours & training   & $.04$ & $.71$ & $.51$ \\
            Ours & validation & $.04$ & $.65$ & $.38$ \\
            \bottomrule
        \end{tabular}
    \end{center}
    \vspace{-1mm}
    \caption{Quantitative evaluation of pose estimation on the PASCAL 3D+ dataset. The used metric is acc${}_{\frac{\pi}{6}}$ in~\cite{tulsiani2015viewpoints}.}
    \label{table:pascal_pose}
    \vspace{-2mm}
\end{table}

\subsection{PASCAL}

We also evaluated our method on PASCAL dataset preprocessed by Tulsiani \etal~\cite{tulsiani2017multi}. This dataset contains three object categories {\it aeroplane}, {\it car}, and {\it chair}, and it is composed of images from PASCAL VOC~\cite{everingham2010pascal}, additional images from ImageNet~\cite{deng2009imagenet}, and their shape and pose annotations provided by PASCAL 3D+~\cite{xiang2014beyond}. Object images were cropped using bounding boxes and resized to the same size. We used ResNet-18 architecture as encoders, and the same decoders as the CIFAR-10 experiments. As feature maps, we used five layers after convolution of pre-trained AlexNet~\cite{krizhevsky2012imagenet}. Though this requires additional supervision provided by ImageNet, this would be replaceable by any kind of self-supervised representation learning methods. Since input images were cropped and resized using bounding boxes, we also cropped and resized rendered images. This reduced the degree of freedom of poses from six to three.

Fig.~\ref{fig:pt_pascal} shows obtained category-specific base shapes and Fig.~\ref{fig:exp_ft_pascal} shows several results of single-view reconstruction by our fully-trained model on the validation set. These results demonstrate that our proposed method works properly on PASCAL dataset. Table~\ref{table:pascal_pose} shows pose estimation accuracy of found best shapes during training on the training set and prediction by the pose estimator on validation set. For {\it car} and {\it chair}, our model found correct shapes during training in the majority of cases, and the pose estimator learned the correspondence of images and poses properly. However, our method cannot find poses of {\it aeroplane} because the ambiguity of poses is very high in this category. Though many of the reconstructed {\it aeroplane} images look plausible at a glance, the accuracy of pose estimation does not correlate with this intuitive evaluation. 

\subsection{Discussion} 

Though we believe that this work is an important step toward 3D understanding without supervision, as the task is very challenging, the accuracy of our method is not very high. In this section, we list our observations in the experiments and possible solutions. Please see the appendix for more qualitative results.

\begin{itemize}
    \vspace{-2mm}
    \setlength\itemsep{0em}
    \item Our method works well for estimating rough shapes and poses, however, it is not very good at reconstructing small details (cf. legs of {\it horses}) and estimating accurate poses in ambiguity (cf. {\it aeroplane}). Also, the intra-class variance of generated shapes is very small. One reason for these problems is that the reconstruction loss using pre-trained image features does not capture category-specific fine-grained features. Actually, a reliable measure of image reconstruction is quite important in self-supervised learning based on render-and-compare loss because reconstruction loss is the only supervision. Incorporating unsupervised learning of keypoints~\cite{suwajanakorn2018discovery} would alleviate this problem. 
    \item We introduced three assumptions in the introduction section. As demonstrated in experiments, these do not prevent learning in {\it car}, {\it horse}, {\it chair}, and {\it aeroplane} categories. However, our proposed method does not work well in {\it cat} and {\it dog} on CIFAR-10 because the deformation of shapes is relatively large. One possible way to learn large deformation would be using videos for training.
    \item We used manually-designed pose distributions in the base shape learning. However, designing them is not straightforward in some categories. For example, the viewpoint distribution of {\it horse} images are far from uniform because there are many photos of zoom up of heads, but fewer photos of tails. How to deal with biased distributions would be an important and interesting problem.
    \item The introduced constraints and regularization may sound too naive and intuitive. Actually, the proposed base shape learning is a bit sensitive to hyperparameters of surface smoothness. This is because of the difficulty to design a robust measure of object naturalness. Learning object naturalness from data~\cite{gwak2017weakly,wu2018learning} would be a promising direction.
\end{itemize}

%% file: 5_conclusion.tex
\section{Conclusion}

In this study, we presented a method to learn single-view reconstruction of the 3D shape, pose, and texture of objects from categorized natural images in a self-supervised manner. The two main techniques were two-stage training to focus on shapes, and inducting strong regularization and constraints to the surface of shapes and background images. Results of experiments on CIFAR-10 and PASCAL confirm the importance of our proposed techniques. In addition, we summarized observations and possible research directions.

%% file: 6_acknowledgment.tex
\subsection*{Acknowledgment}
\addcontentsline{toc}{section}{Acknowledgment}
\vspace{-1mm}

\small{This work was partially supported by JST CREST Grant Number JPMJCR1403, and partially supported by JSPS KAKENHI Grant Number JP19H01115.}

%% file: 7_appendix.tex
\section{Appendix}

\begin{figure*}[t]
    \begin{center}
        \includegraphics[width=15mm]{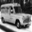}
        \includegraphics[width=15mm]{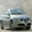}
        \includegraphics[width=15mm]{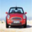}
        \includegraphics[width=15mm]{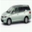}
        \includegraphics[width=15mm]{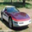}
        \includegraphics[width=15mm]{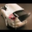}
        \includegraphics[width=15mm]{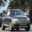}
        \includegraphics[width=15mm]{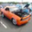}
        \includegraphics[width=15mm]{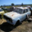}
        \includegraphics[width=15mm]{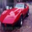}
        \includegraphics[width=15mm]{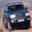}
        \includegraphics[width=15mm]{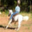}
        \includegraphics[width=15mm]{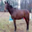}
        \includegraphics[width=15mm]{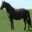}
        \includegraphics[width=15mm]{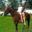}
        \includegraphics[width=15mm]{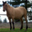}
        \includegraphics[width=15mm]{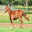}
        \includegraphics[width=15mm]{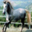}
        \includegraphics[width=15mm]{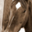}
        \includegraphics[width=15mm]{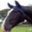}
        \includegraphics[width=15mm]{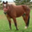}
        \includegraphics[width=15mm]{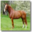}
        \includegraphics[width=15mm]{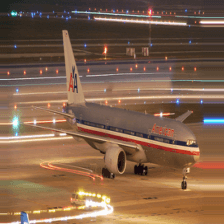}
        \includegraphics[width=15mm]{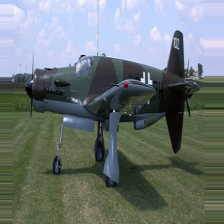}
        \includegraphics[width=15mm]{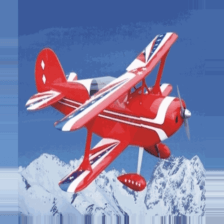}
        \includegraphics[width=15mm]{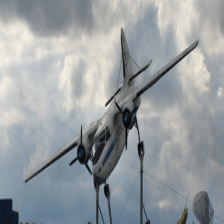}
        \includegraphics[width=15mm]{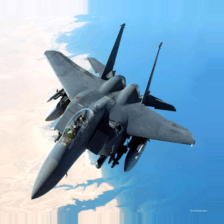}
        \includegraphics[width=15mm]{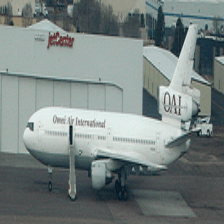}
        \includegraphics[width=15mm]{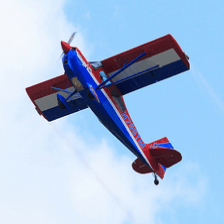}
        \includegraphics[width=15mm]{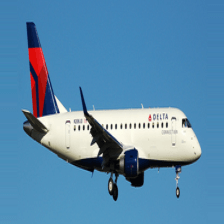}
        \includegraphics[width=15mm]{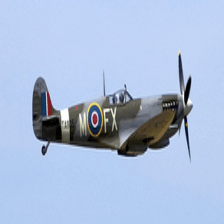}
        \includegraphics[width=15mm]{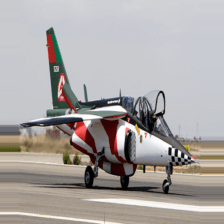}
        \includegraphics[width=15mm]{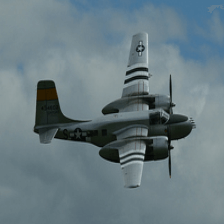}
        \includegraphics[width=15mm]{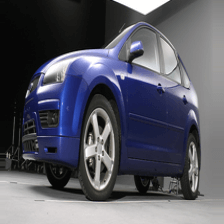}
        \includegraphics[width=15mm]{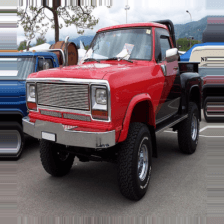}
        \includegraphics[width=15mm]{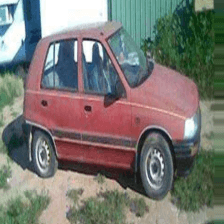}
        \includegraphics[width=15mm]{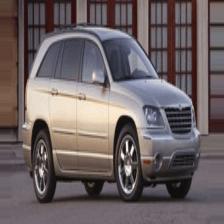}
        \includegraphics[width=15mm]{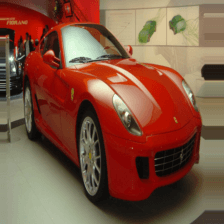}
        \includegraphics[width=15mm]{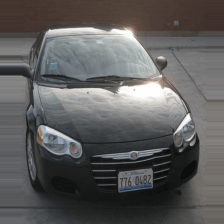}
        \includegraphics[width=15mm]{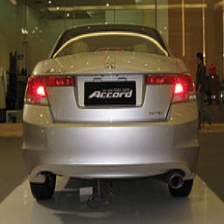}
        \includegraphics[width=15mm]{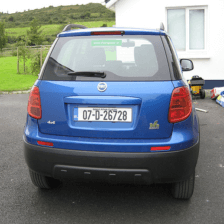}
        \includegraphics[width=15mm]{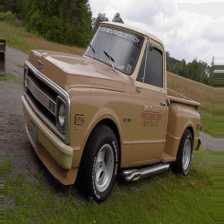}
        \includegraphics[width=15mm]{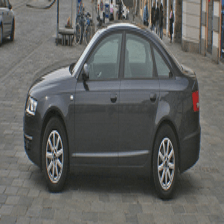}
        \includegraphics[width=15mm]{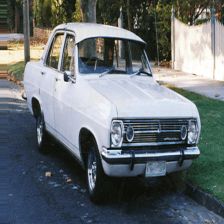}
        \includegraphics[width=15mm]{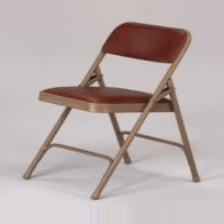}
        \includegraphics[width=15mm]{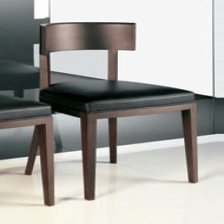}
        \includegraphics[width=15mm]{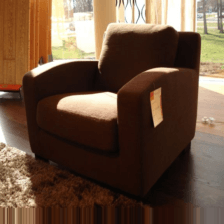}
        \includegraphics[width=15mm]{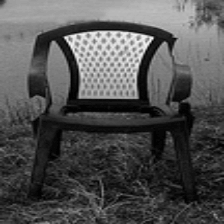}
        \includegraphics[width=15mm]{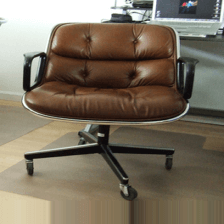}
        \includegraphics[width=15mm]{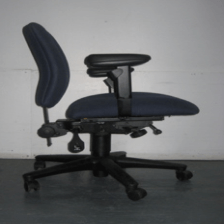}
        \includegraphics[width=15mm]{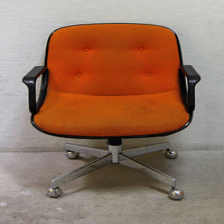}
        \includegraphics[width=15mm]{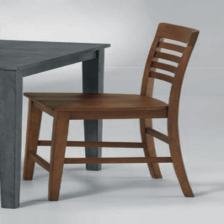}
        \includegraphics[width=15mm]{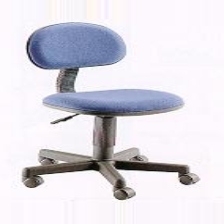}
        \includegraphics[width=15mm]{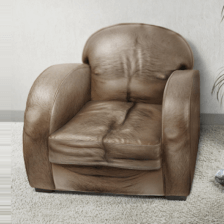}
        \includegraphics[width=15mm]{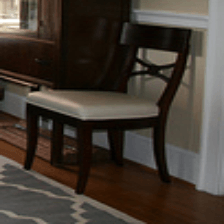}
    \end{center}
    \caption{Randomly selected images in our training dataset. From top to bottom: CIFAR-10 {\it car}, CIFAR-10 {\it horse}, PASCAL {\it aeroplane}, PASCAL {\it car}, and PASCAL {\it chair}. }
    \label{figure:training_data}
\end{figure*}

\begin{figure*}[t!]
    \begin{center}
        \parbox{28mm}{\raisebox{16mm}{\parbox{28mm}{(a) Ours on \\ PASCAL}}}
        \includegraphics[width=15mm]{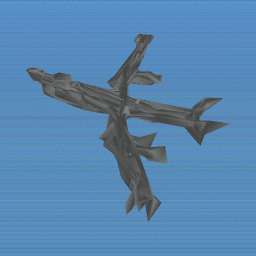}
        \includegraphics[width=15mm]{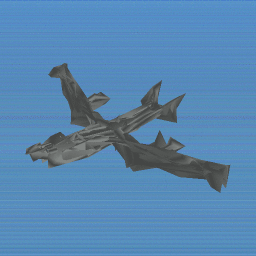}
        \includegraphics[width=15mm]{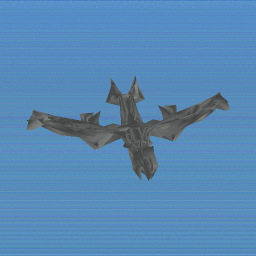}
        \includegraphics[width=15mm]{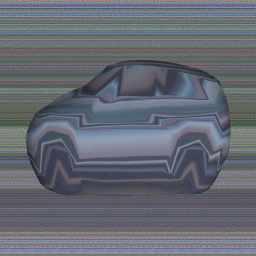}
        \includegraphics[width=15mm]{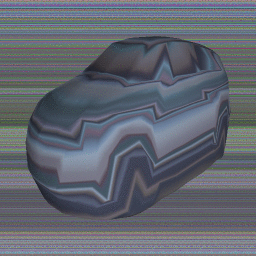}
        \includegraphics[width=15mm]{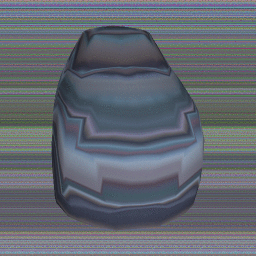}
        \includegraphics[width=15mm]{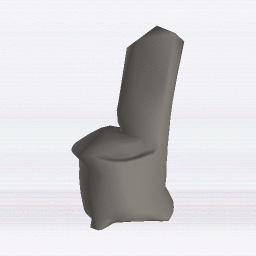}
        \includegraphics[width=15mm]{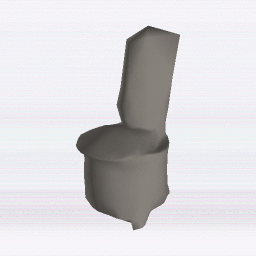}
        \includegraphics[width=15mm]{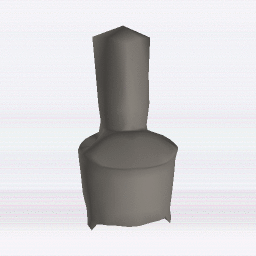} \vspace{-9.0mm} \\
        \parbox{28mm}{\raisebox{16mm}{\parbox{28mm}{(b) HoloGAN on \\ PASCAL}}}
        \includegraphics[width=15mm]{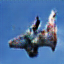}
        \includegraphics[width=15mm]{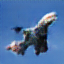}
        \includegraphics[width=15mm]{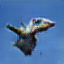}
        \includegraphics[width=15mm]{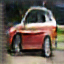}
        \includegraphics[width=15mm]{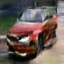}
        \includegraphics[width=15mm]{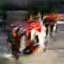}
        \includegraphics[width=15mm]{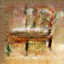}
        \includegraphics[width=15mm]{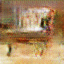}
        \includegraphics[width=15mm]{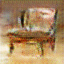} \vspace{-9.0mm} \\
    \end{center}
    \caption{Comparison of ours with HoloGAN on PASCAL dataset. Images are rendered at 50 degree intervals.}
    \label{fig:append_hologan_pascal}
\end{figure*}

\subsection{Implementation details}

\subsubsection{Smoothness regularization}
As described in Section~\ref{sec:method_pt_shape}, we minimize graph Laplacian of a mesh, which represents approximated mean curvature at each vertex~\cite{taubin1995signal}, and angles between two neighboring triangle polygons, which implies smoothness at each edge. These regularizations are represented by a loss term $\mathcal{L}_s$. For a mesh of $N_v$ vertices, let $L \in \mathcal{R}^{N_v \times N_v}$ be a Laplacian matrix of vertices $v \in \mathcal{R}^{N_v \times 3}$ and let $t_{l} \in \mathcal{R}^{+}$ be a hyper parameter that controls tolerance. We minimize $\mathcal{L}_{g_2} = \sum_i \|L_i v\|_2^2$ and $\mathcal{L}_{g_1} = \max ((\sum_i \|L_i v\|_2^1) - t_l, 0)^2$. Similarly, let $\theta_{i,j}$ be the angle between normal vectors of two neighboring triangle polygons $f_i$ and $f_j$ and let $t_{a} \in \mathcal{R}^{+}$ be a hyper parameter, we minimize $\mathcal{L}_{a_2} = \sum_{i,j} \theta_{i, j}^2$ and $\mathcal{L}_{a_1} = \max (\| \sum_{i, j} \theta_{i, j} \| - t_a, 0)^2$. With weighting parameters $\lambda_{g_1}, \lambda_{g_2}, \lambda_{a_1}, \lambda_{a_2}$, the sum of these regularization terms is 
\begin{equation}
    \mathcal{L}_s = \lambda_{g_1} \mathcal{L}_{g_1} + \lambda_{g_2} \mathcal{L}_{g_2} + \lambda_{a_1} \mathcal{L}_{a_1} + \lambda_{a_2} \mathcal{L}_{a_2}.
\end{equation}

\subsubsection{Initial object dimensions}
As described in Section~\ref{sec:method_pt_shape}, we manually give a rough shape in base shape learning by setting initial dimensions of a predefined sphere to alleviate the difficulty of learning aspect ratio of objects. Table~\ref{table:append_shape_dimensions} shows the setting. Our method worked well even though the specified values were very rough.

\subsubsection{Viewpoint distribution}
As described in Section~\ref{sec:method_pt_pose}, we use predefined viewpoint distributions in learning base shapes. Table~\ref{table:append_viewpoints} shows the setting.

\subsubsection{Reconstruction loss}
As described in Section~\ref{sec:method_pt_reconstruction}, we use feature matching~\cite{salimans2016improved} and the Chamfer distance of real and generated image minibatches to compute a reconstruction loss $\mathcal{L}_{\text{rec}}$ in the base shape learning. Let $H, W$ be the height and width of images assuming all images are resized to the same size. Given $N_{c}$-channel feature maps of $N_{b}$ real images $f \in \mathcal{R}^{N_b \times N_{c} \times H \times W}$ and $N_{b}$ generated images $\hat{f}$, we define the distance of two image features $f_{i}, \hat{f}_{i'}$ as
\begin{equation}
   \mathcal{D} (f_{i}, \hat{f_{i'}}) = \frac{1}{HW} \sum_{k=1}^{H}\sum_{l=1}^{W} \sqrt{\sum_{j=1}^{N_c} (\hat{f}_{{i'}jkl} -  f_{{i}jkl})^2}.
   \label{eq:distance}
\end{equation}
Using this distance, feature matching loss of $f$ and $\hat{f}$ is defined as 
\begin{equation}
   \mathcal{L}_{\text{fm}} = \mathcal{D} (\sum_{i=1}^{N_b} \frac{f_i}{N_b}, \sum_{i=1}^{N_b} \frac{\hat{f}_{i}}{N_b}),
   \label{eq:feature_matching}
\end{equation}
and Chamfer distance is defined as
\begin{equation}
   \mathcal{L}_{\text{cd}} = \frac{1}{N_n} (\sum_{i=1}^{N_b} \min_{i'} \mathcal{D} (f_i, \hat{f}_{i'}) + \sum_{i'=1}^{N_b} \min_{i} \mathcal{D} (\hat{f}_{i'}, f_i)).
\end{equation}
With a hyper parameter $\lambda_{\text{rec}}$, we define reconstruction loss as $\mathcal{L}_{\text{rec}} = \lambda_{\text{rec}} \mathcal{L}_{\text{cd}} + (1 - \lambda_{\text{rec}}) \mathcal{L}_{\text{fm}}$. We set $\lambda_{\text{rec}}$ to $0.9, 0.3, 0.3, 0.9$, and $0.8$ for CIFAR-10 {\it car}, CIFAR-10 {\it horse}, PASCAL {\it aeroplane}, PASCAL {\it car}, and PASCAL {\it chair} respectively.

\subsubsection{Optimization}
In all experiments, we used Adam optimizer with $\alpha=0.0001$, $\beta_1=0.5$, and $\beta_2=0.99$. In base shape learning, batch size is set to $64$, and the number of training iterations of the base shape learning is set to $1200, 300, 200, 800$, and $200$ for CIFAR-10 {\it car}, CIFAR-10 {\it horse}, PASCAL {\it aeroplane}, PASCAL {\it car}, and PASCAL {\it chair} respectively. In full model training, the number of iterations is set to $10000$ in all categories.

\begin{table}[t]
    \begin{center}
        \small
        \begin{tabular}{lccc}
            \toprule
            Category & Width & Height & Depth \\
            \hline
            {\it car}       & 0.5 & 0.5 & 1.0 \\
            {\it horse}     & 1.0 & 1.0 & 1.0 \\
            {\it aeroplane} & 1.0 & 0.5 & 1.0 \\
            {\it chair}     & 1.0 & 1.0 & 1.0 \\
            \bottomrule
        \end{tabular}
    \end{center}
    \caption{Initial dimensions in base shape learning.}
    \label{table:append_shape_dimensions}
\end{table}

\begin{table}[t]
    \begin{center}
        \small
        \begin{tabular}{lll}
            \toprule
            Category & Elevation & Azimuth \\
            \hline
            {\it car}       & $\text{Uniform}$ (0, 30) & $\text{Uniform} (0, 360) $ \\
            {\it horse}     & $\text{Uniform}$ (-10, 10) & $\text{Beta} (1.5, 1.5) * 180 $ \\
            {\it aeroplane} & $\text{Uniform}$ (-60, 60) & $\text{Beta} (1.5, 1.5) * 180 $ \\
            {\it chair}     & $\text{Uniform}$ (0, 30) & $\text{Uniform} (0, 360) $ \\
            \bottomrule
        \end{tabular}
    \end{center}
    \caption{Predefined distribution of viewpoints. When azimuth is zero, the camera is located in front of the object.}
    \label{table:append_viewpoints}
\end{table}

\subsection{Additional experimental results}

\subsubsection{Training data}
Fig.~\ref{figure:training_data} shows randomly selected training samples from our experiments. Different from commonly used datasets such as ShapeNet, foregrounds and backgrounds are not separated, viewpoints are unknown, objects have various shapes and sizes, and they locate and rotate freely. Though learning about 3D from these datasets is very challenging due to these characteristics, we think that trying this is a necessary step toward fundamental 3D understanding in machines.

\subsubsection{Comparison with HoloGAN on PASCAL}
We showed comparison between ours and HoloGAN on CIFAR-10 in Fig.~\ref{fig:hologan_cifar10}. For this comparison, we used codes provided by the authors. Specifically, we used default settings for CelebA dataset except for elevation and azimuth, which are set to the values for the cars included in the paper. In addition, we show comparison on PASCAL dataset in Fig.~\ref{fig:append_hologan_pascal}. Though images generated by HoloGAN are improved, they still lack consistency from multiple viewpoints. Especially, this method seems not good at generating front images of cars.

\subsubsection{Randomly selected results}
\begin{figure*}[t]
    \begin{center}
        \includegraphics[width=\linewidth]{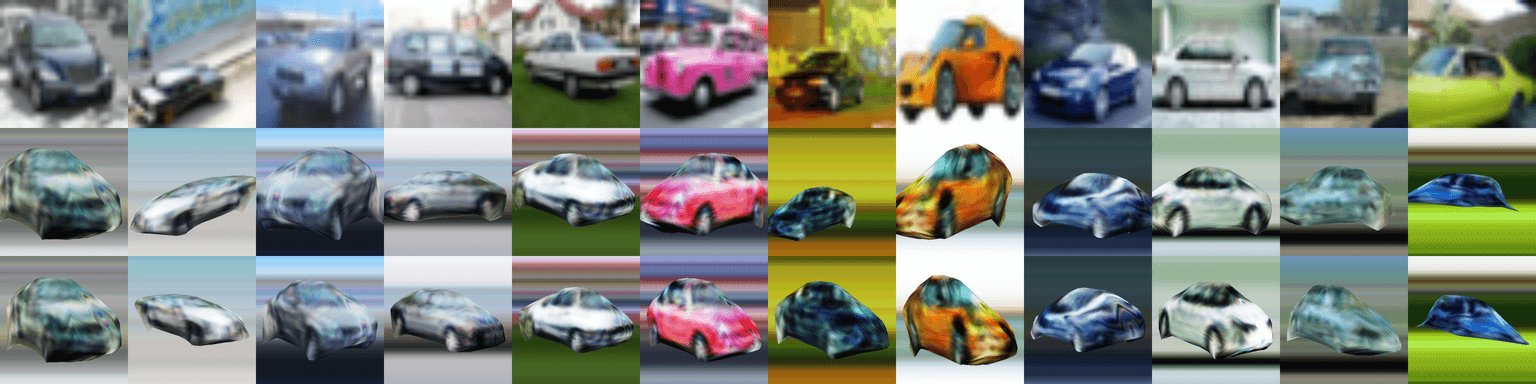}  \\
        \vspace{1mm} 
        \includegraphics[width=\linewidth]{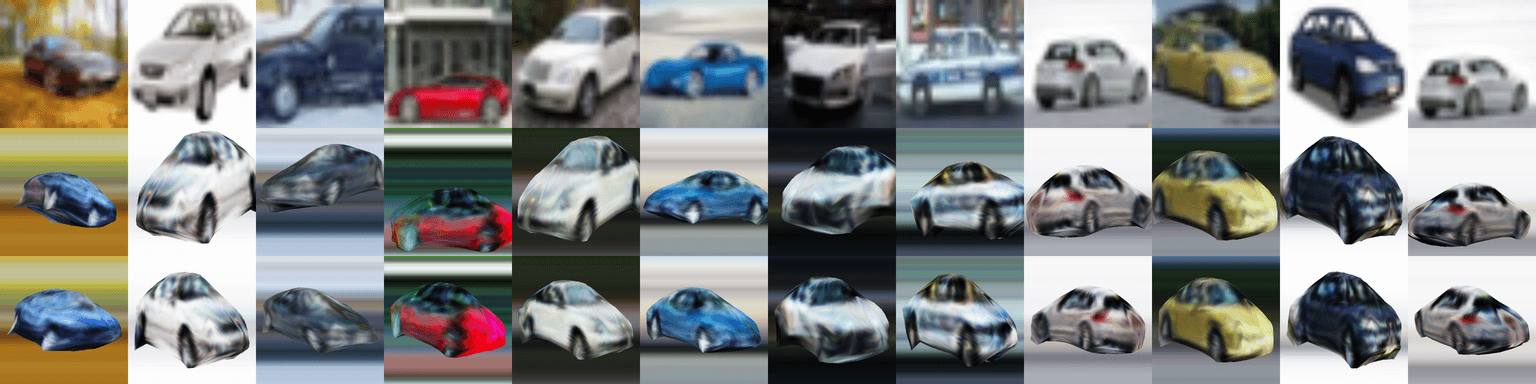}  \\
    \end{center}
    \caption{Randomly selected results on CIFAR-10 {\it car} training set.}
    \label{fig:append_random_1}
\end{figure*}

\begin{figure*}[t]
    \begin{center}
        \includegraphics[width=\linewidth]{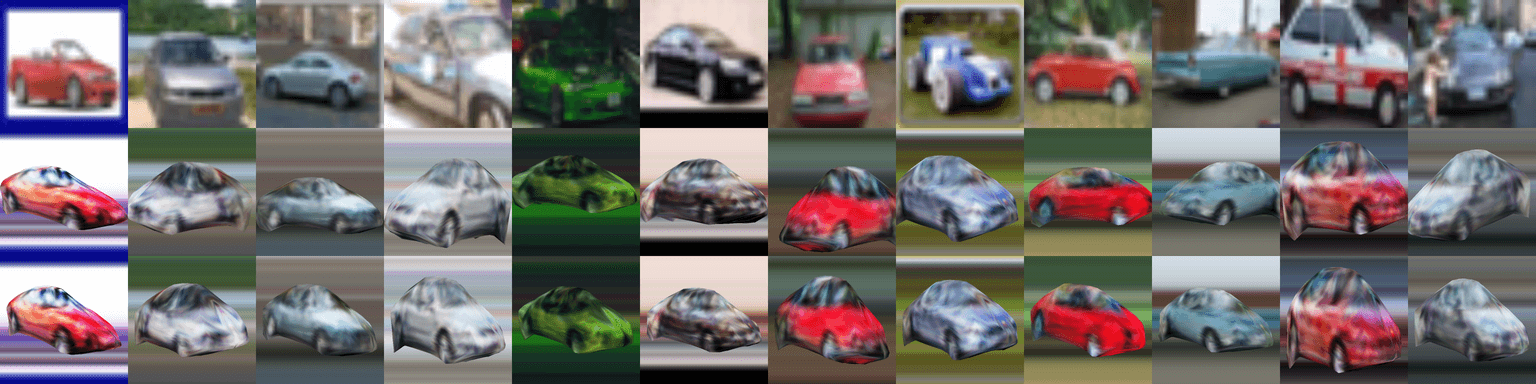}  \\
        \vspace{1mm} 
        \includegraphics[width=\linewidth]{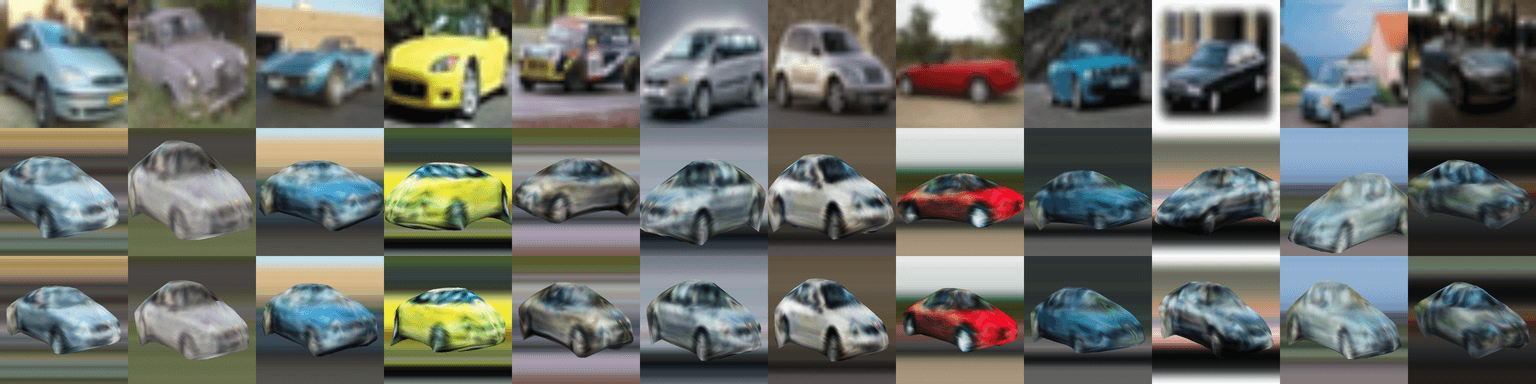}  \\
    \end{center}
    \caption{Randomly selected results on CIFAR-10 {\it car} validation set.}
    \label{fig:append_random_2}
\end{figure*}

\begin{figure*}[t]
    \begin{center}
        \includegraphics[width=\linewidth]{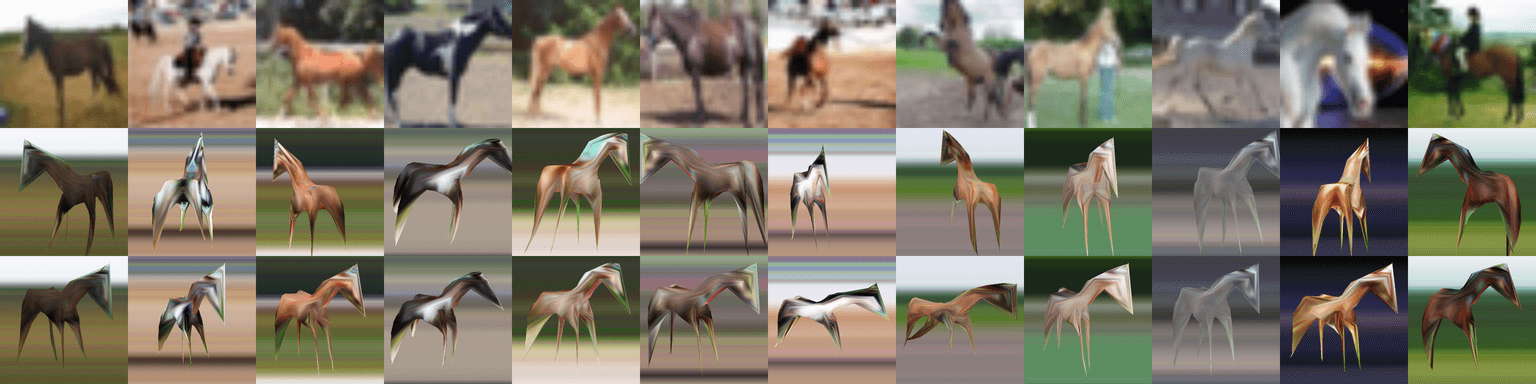}  \\
        \vspace{1mm} 
        \includegraphics[width=\linewidth]{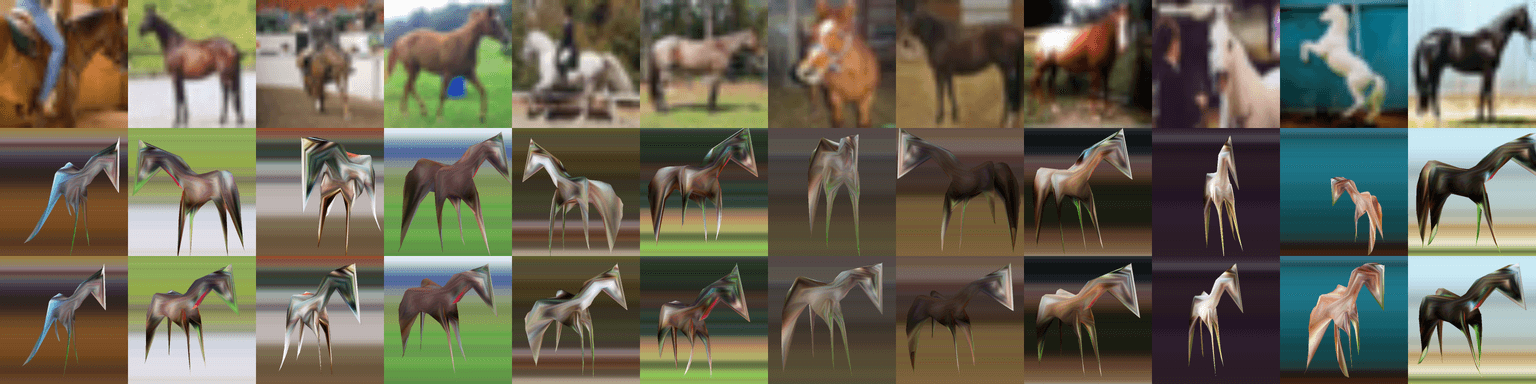}  \\
    \end{center}
    \caption{Randomly selected results on CIFAR-10 {\it horse} training set.}
    \label{fig:append_random_3}
\end{figure*}

\begin{figure*}[t]
    \begin{center}
        \includegraphics[width=\linewidth]{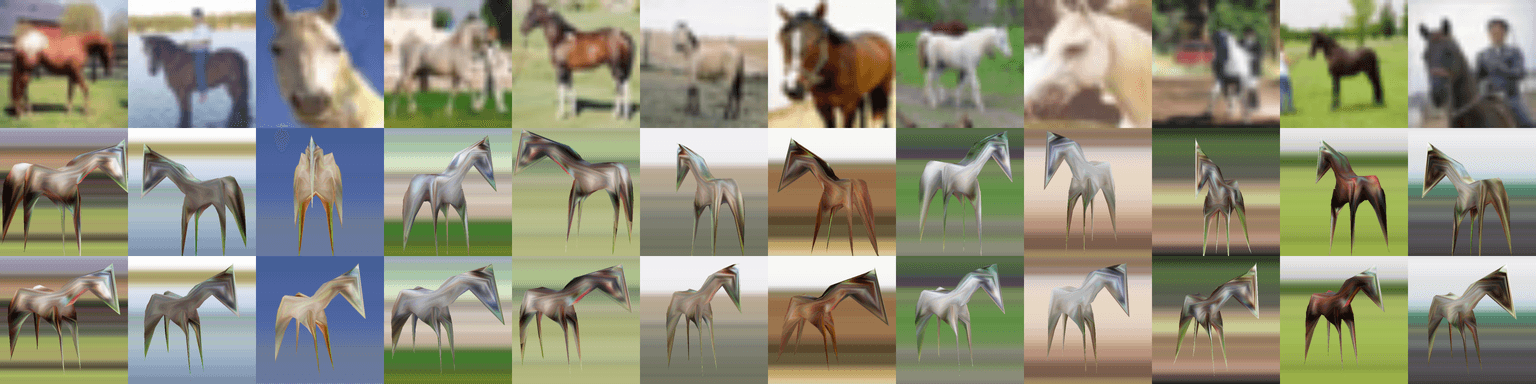}  \\
        \vspace{1mm} 
        \includegraphics[width=\linewidth]{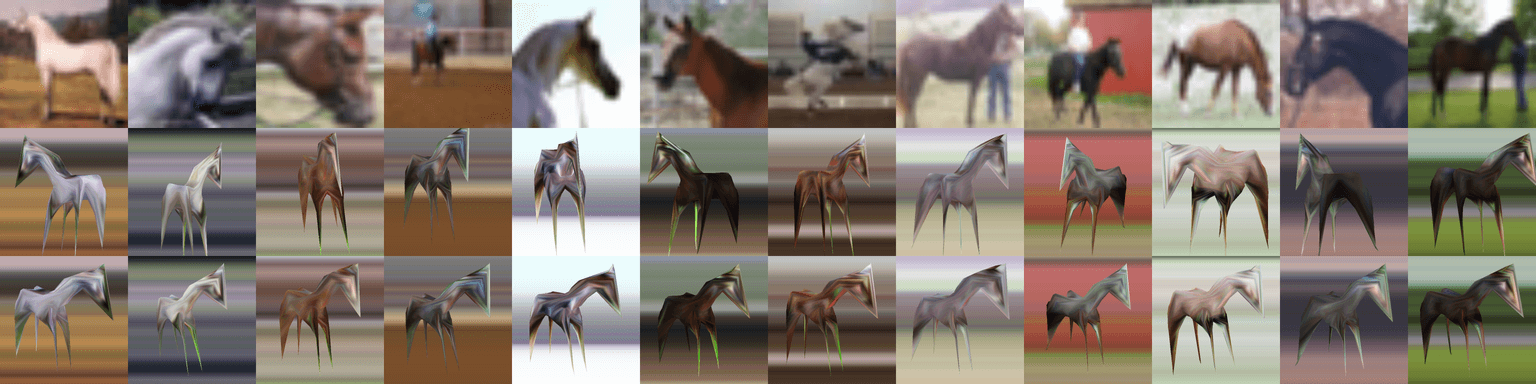}  \\
    \end{center}
    \caption{Randomly selected results on CIFAR-10 {\it horse} validation set.}
    \label{fig:append_random_4}
\end{figure*}

\begin{figure*}[t]
    \begin{center}
        \includegraphics[width=\linewidth]{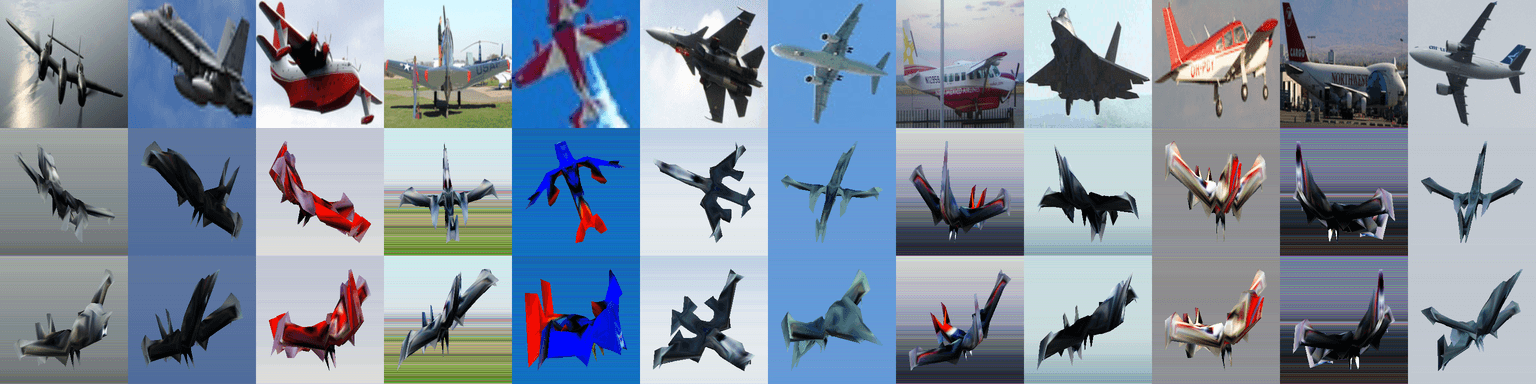}  \\
        \vspace{1mm} 
        \includegraphics[width=\linewidth]{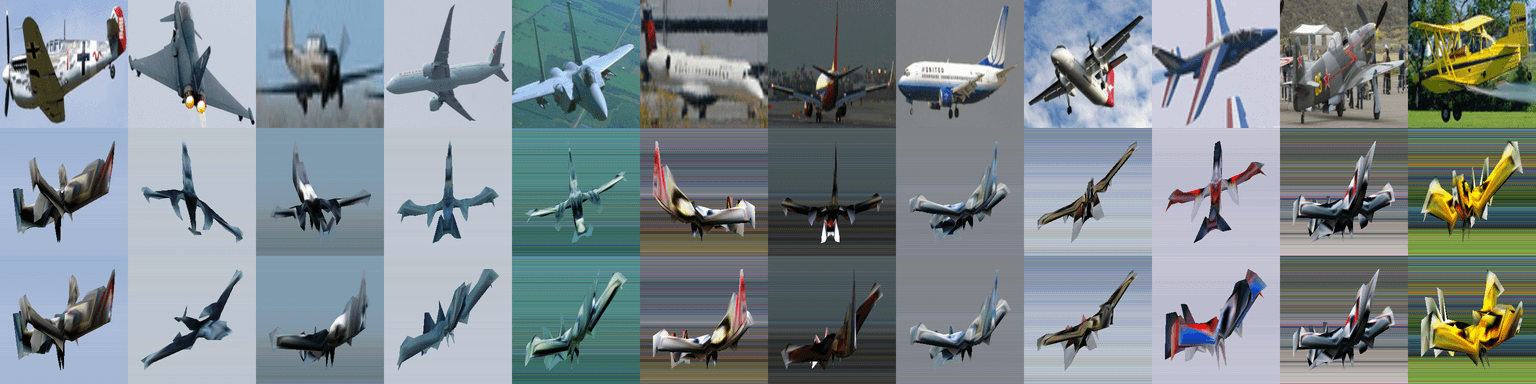}  \\
    \end{center}
    \caption{Randomly selected results on PASCAL {\it aeroplane} training set.}
    \label{fig:append_random_5}
\end{figure*}

\begin{figure*}[t]
    \begin{center}
        \includegraphics[width=\linewidth]{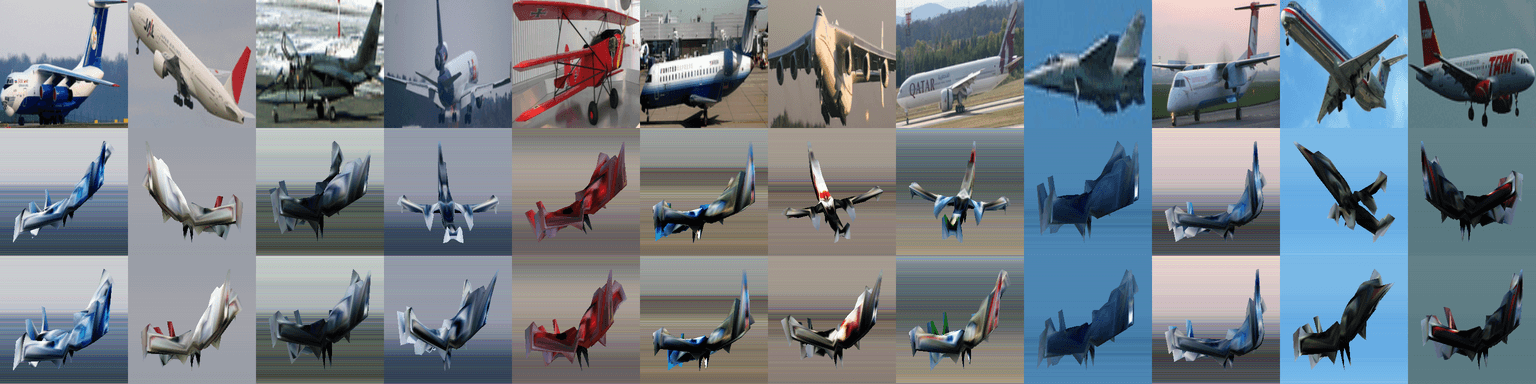}  \\
        \vspace{1mm} 
        \includegraphics[width=\linewidth]{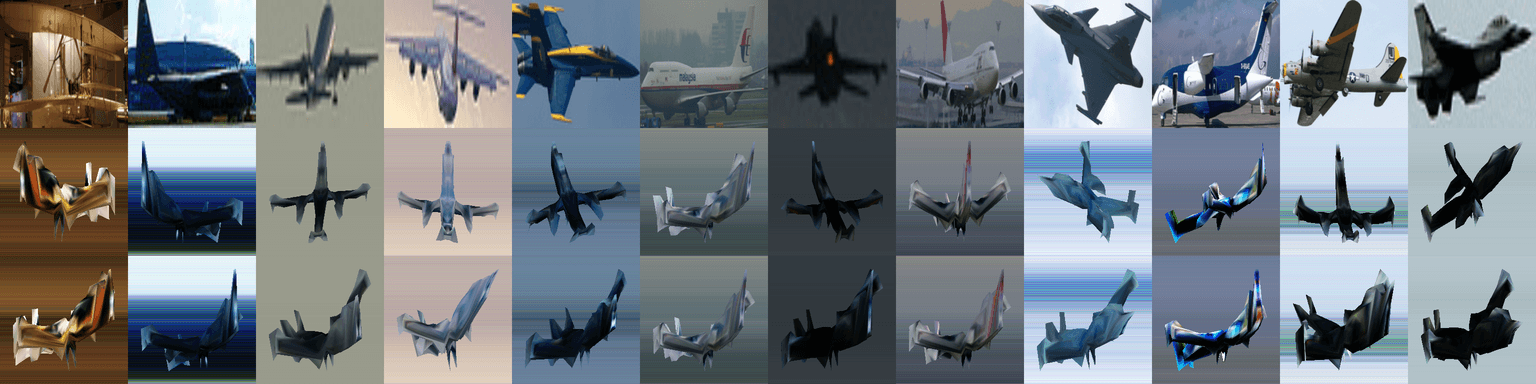}  \\
    \end{center}
    \caption{Randomly selected results on PASCAL {\it aeroplane} validation set.}
    \label{fig:append_random_6}
\end{figure*}

\begin{figure*}[t]
    \begin{center}
        \includegraphics[width=\linewidth]{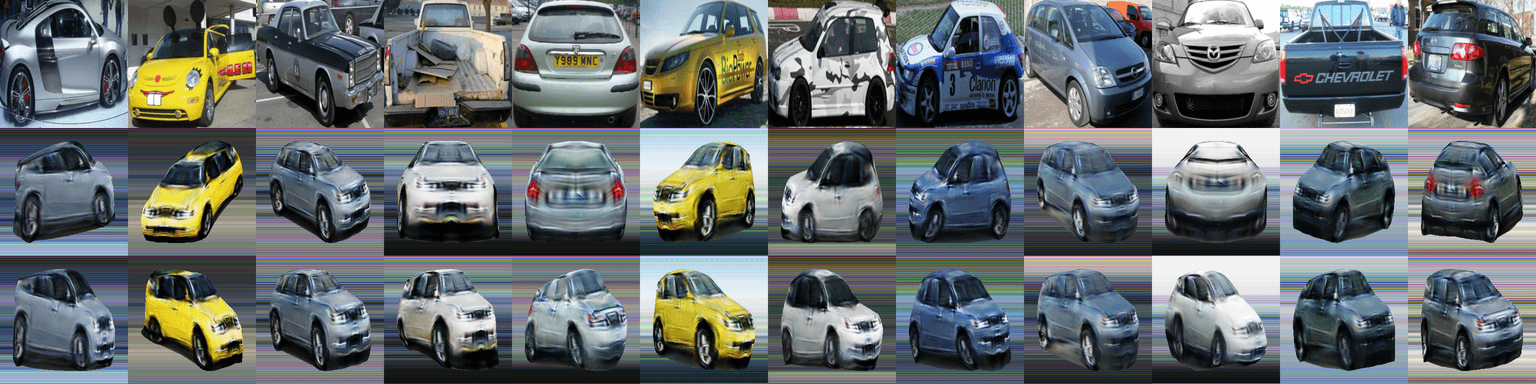}  \\
        \vspace{1mm} 
        \includegraphics[width=\linewidth]{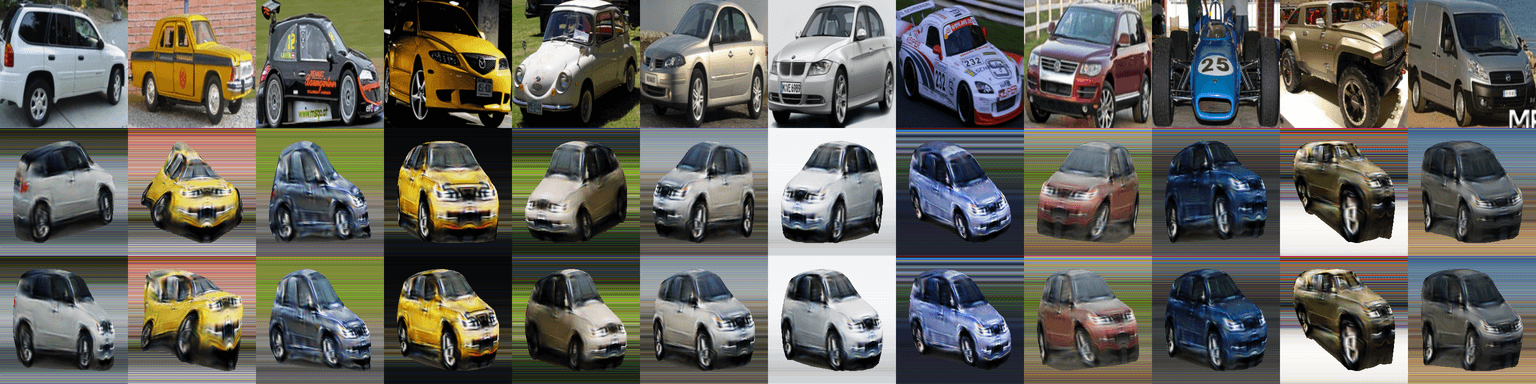}  \\
    \end{center}
    \caption{Randomly selected results on PASCAL {\it car} training set.}
    \label{fig:append_random_7}
\end{figure*}

\begin{figure*}[t]
    \begin{center}
        \includegraphics[width=\linewidth]{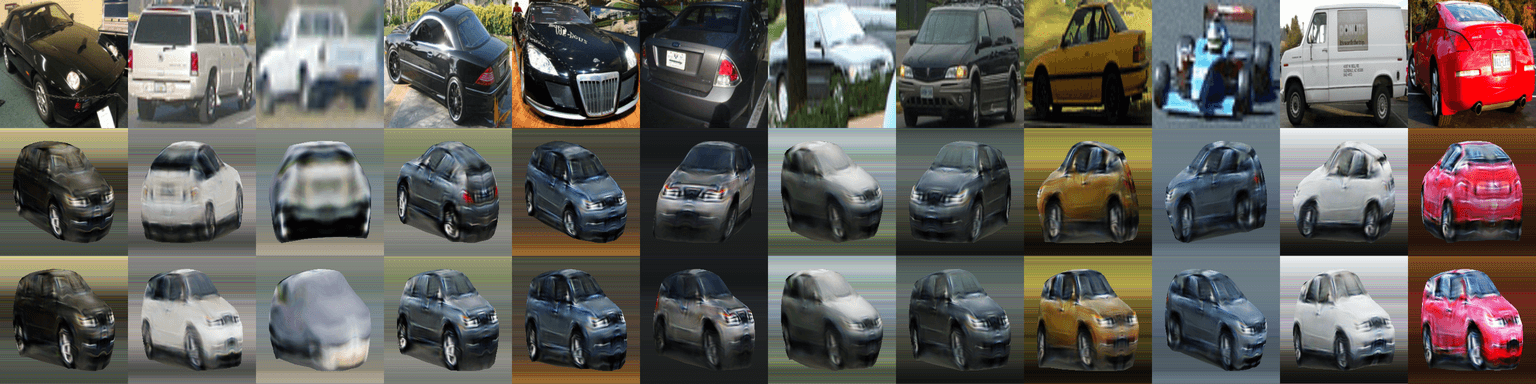}  \\
        \vspace{1mm} 
        \includegraphics[width=\linewidth]{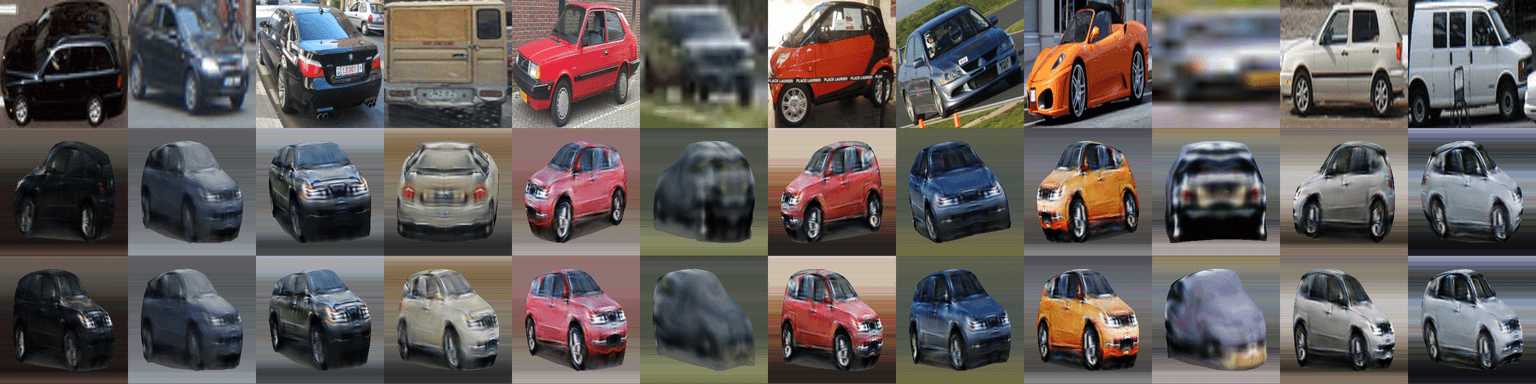}  \\
    \end{center}
    \caption{Randomly selected results on PASCAL {\it car} validation set.}
    \label{fig:append_random_8}
\end{figure*}

\begin{figure*}[t]
    \begin{center}
        \includegraphics[width=\linewidth]{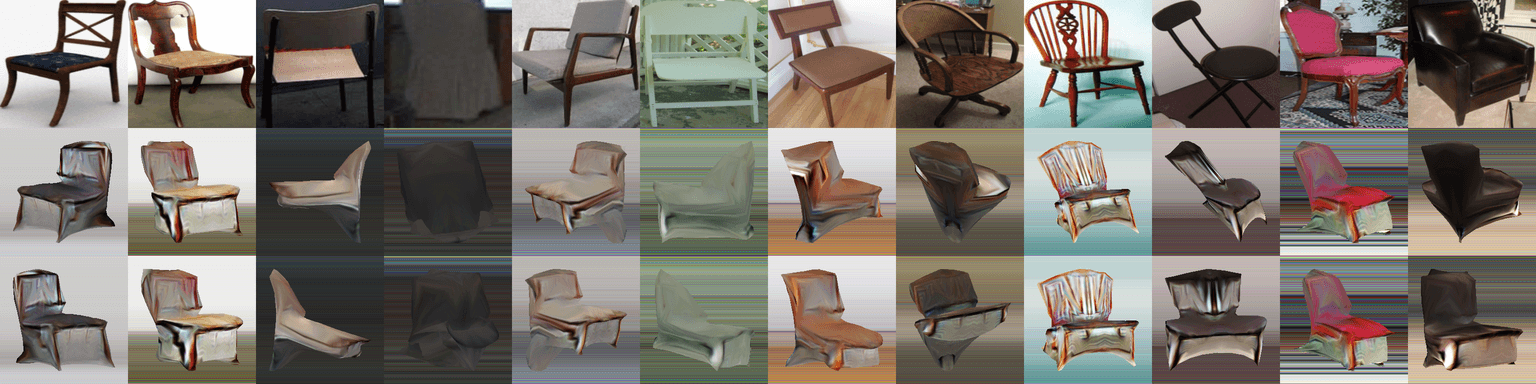}  \\
        \vspace{1mm} 
        \includegraphics[width=\linewidth]{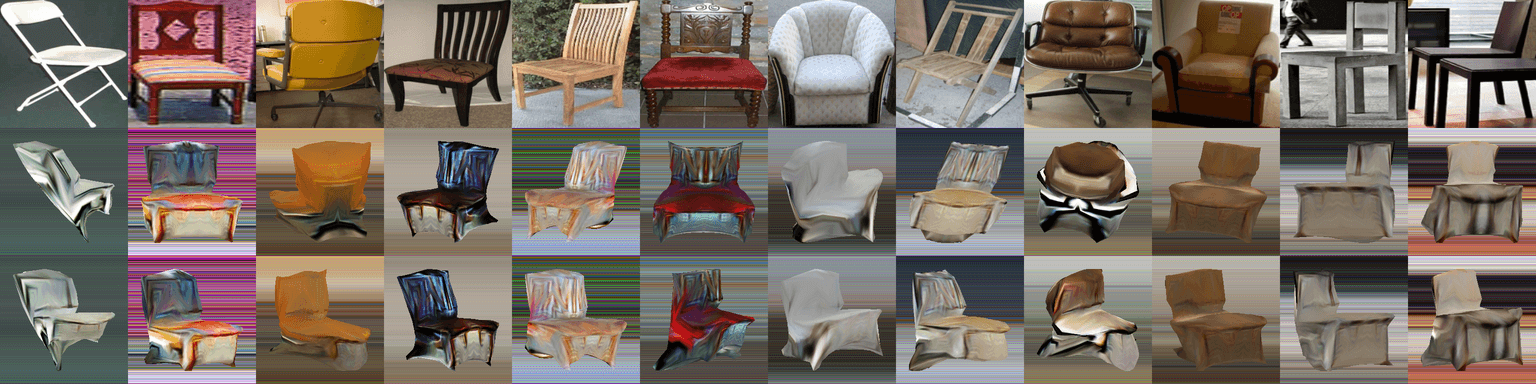}  \\
    \end{center}
    \caption{Randomly selected results on PASCAL {\it chair} training set.}
    \label{fig:append_random_9}
\end{figure*}

\begin{figure*}[t]
    \begin{center}
        \includegraphics[width=\linewidth]{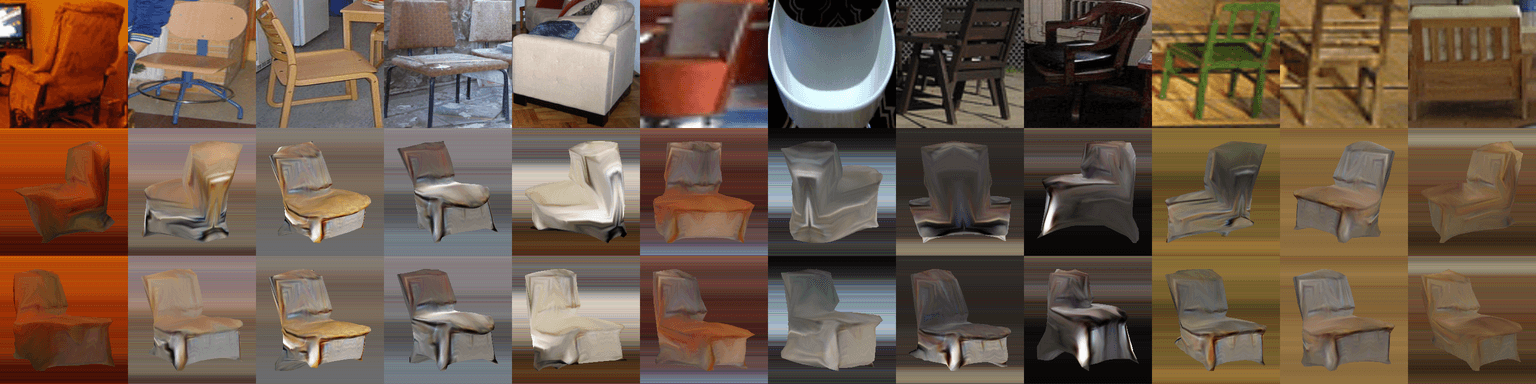}  \\
        \vspace{1mm} 
        \includegraphics[width=\linewidth]{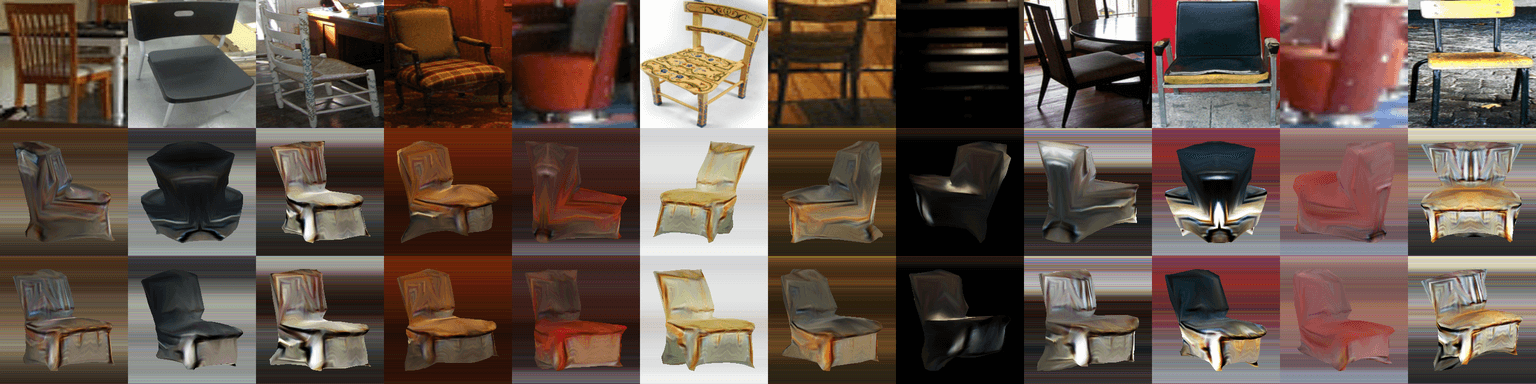}  \\
    \end{center}
    \caption{Randomly selected results on PASCAL {\it chair} validation set.}
    \label{fig:append_random_10}
\end{figure*}

In Fig.~\ref{fig:exp_ft} and Fig.~\ref{fig:exp_ft_pascal}, selected results on CIFAR-10 and PASCAL datasets are shown. For further qualitative evaluation, randomly selected results are shown in Fig.~\ref{fig:append_random_1} to Fig.~\ref{fig:append_random_10}. Input images are shown in the top rows, reconstructed images using estimated shapes, poses, textures, and backgrounds are shown in the middle rows, and reconstructed images using a fixed viewpoint are shown in the bottom rows. For training set, the best shapes and viewpoints found during training are used, and for validation set, predicted shapes and viewpoints by the shape estimator and viewpoint estimator are used. Photometric fine-tuning are used for only validation set.

\clearpage
